\documentclass[twoside,leqno,twocolumn]{article}

\usepackage[letterpaper]{geometry}

\usepackage{ltexpprt}
\usepackage{hyperref}

\usepackage{amsmath, mathtools, amssymb, color}
\usepackage{comment}
\usepackage{tikz}
\usepackage{booktabs}

\usepackage{subfig}
\usepackage{graphicx}

\usepackage{algorithm}
\usepackage{algpseudocode}

\begin{document}

\newcommand\relatedversion{}

\title{\Large An Interpretable Measure for Quantifying Predictive Dependence between Continuous Random Variables} 
%\author{Anonymous Submission}
\author{Renato Assun\c{c}\~{a}o\footnotemark[2]~\thanks{ESRI Inc.}
\and Fl\'{a}vio Figueiredo\thanks{Computer Science Department, DCC-UFMG, Brazil.}
\and Francisco N. Tinoco Junior\footnotemark[2]
\and Leo M. de S\'{a}-Freire\footnotemark[2]
\and F\'{a}bio Silva\thanks{Federal Center for Technological Education, CEFET, Brazil.}
}

\date{}

\maketitle

% Default Copyright Statement
%\fancyfoot[R]{\scriptsize{Copyright \textcopyright\ 2025 by SIAM\\
% Unauthorized reproduction of this article is prohibited}}

%\pagenumbering{arabic}
%\setcounter{page}{1}%Leave this line commented out.

\begin{abstract} \small\baselineskip=9pt A fundamental task in statistical learning is quantifying the joint dependence or association between two continuous random variables. We introduce a novel, fully non-parametric measure that assesses the degree of association between continuous variables $X$ and $Y$, capable of capturing a wide range of relationships, including non-functional ones. A key advantage of this measure is its interpretability: it quantifies the expected relative loss in predictive accuracy when the distribution of $X$ is ignored in predicting $Y$. This measure is bounded within the interval [0,1] and is equal to zero if and only if $X$ and $Y$ are independent. We evaluate the performance of our measure on over 90,000 real and synthetic datasets, benchmarking it against leading alternatives. Our results demonstrate that the proposed measure provides valuable insights into underlying relationships, particularly in cases where existing methods fail to capture important dependencies. \end{abstract}

\section{Introduction}\label{sec:intro}

\begin{figure}[t!]
    \centering
    \begin{minipage}{\columnwidth}
    \small
\begin{tabular}{ccccccc|c}\\ \hline
\hline
\multicolumn{ 3}{c}{Rows}&\multicolumn{ 4}{c}{Columns (Variable $X$)}&\\ \cline{4-8}
\multicolumn{ 3}{c}{(Variable $Y$) }&1 &  2 &  $\cdots$ &  c &Total\\ \hline
&\multicolumn{ 2}{c}{1}&$n_{11}$    &   $n_{12}$&   $\cdots$        &$n_{1c}$&$n_{1.}$      \\
&\multicolumn{ 2}{c}{2}&$n_{21}$&   $n_{22}$&   $\cdots$        &   $n_{2c}$&$n_{2.}$\\
&\multicolumn{ 2}{c}{$\vdots$}&$\vdots$ &$\vdots$&$\ddots$& $\vdots$&$\vdots$\\
&\multicolumn{ 2}{c}{r}&$n_{r1}$    &   $n_{r2}$    &   $\cdots$    &   $n_{rc}$&$n_{r.}$       \\ \hline
&\multicolumn{2}{c}{Total}&$n_{.1}$&$n_{.2}$&$\cdots$&$n_{.c}$&$ n_{\cdot \cdot}$ \\ \hline
\hline
\end{tabular}
    \end{minipage}\hfill
    \begin{minipage}{0.40\textwidth}
        \centering
        \includegraphics[width=\columnwidth]{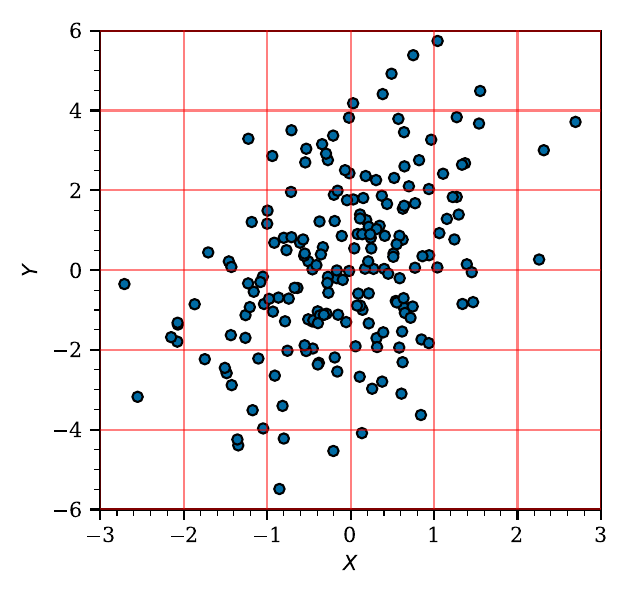}
        % \caption{$dt =$}
        % \label{fig:prob1_6_1}
    \end{minipage}
    \caption{Left: Illustrative contingency table. Right: Regular grid on top of a scatterplot of a random sample $(x_k, y_k)$, $k=1, \ldots, N$ of the random vector $(X,Y)$ with joint density $f_{XY}(x,y)$.}
    \label{fig:ContabAndGrid}
\end{figure}

One of the foundational tasks in data science involves quantifying the degree of dependence or association between two or more random variables, denoted as $X$ and $Y$ \cite{tjostheim2022statistical, reimherr2013quantifying}.
We are inspired by the association measure $\tau_b$ by Goodman and Kruskal \cite{goodman1954measures} for two categorical random variables. This measure is calculated based on discrete data organized in a contingency table (see Figure \ref{fig:ContabAndGrid}). It calculates the reduction in the probability of incorrectly predicting the category of the $Y$ variable when information about the category of $X$ is available. $\tau_b$ is one of the few dependence measures with a clear empirical interpretation~\cite{reimherr2013quantifying}. This clarity of interpretation is the main appeal for us when generalizing the measure to continuous random variables.

% \begin{figure}
%     \centering
%     \includegraphics[width=0.4\linewidth]{figures/ContingencyTable.png}
%     \includegraphics[width=0.4\linewidth]{figures/RegularGrid.png}
%     \caption{Left: Illustrative contingency table. Right: Regular grid on top of a scatterplot of a random sample $(x_k, y_k)$, $k=1, \ldots, N$ of the random vector $(X,Y)$ with joint density $f_{XY}(x,y)$.}
%     \label{fig:ContabAndGrid}
% \end{figure}

We begin with a regular grid superimposed on a scatterplot of independent and identically distributed random points drawn from a \textit{continuous} random vector distribution. This grid will be taken to the limit as the grid spacing approaches zero in our measure and, consequently, will not appear in the final measure definition.
We made two fundamental changes from the original proposal in \cite{goodman1954measures} to ensure meaningful results. As detailed in Section \ref{sec:new-measure}, Goodman and Kruskal considered random assignments of the data into table categories and counted the number of instances where the true data categories matched the randomly assigned ones. In contrast, we consider the continuous case by using the expected \textit{rate} of \textit{correct} random assignments, and taking the asymptotic limit as the sample size approaches infinity and the regular grid spacing approaches zero. We find that a modified version of Goodman and Kruskal's measure can be expressed in terms of the continuous marginal and conditional densities of $X$ and $Y$. The resulting measure is linked to the quadratic R\'{e}nyi entropy \cite{renyi1961measures, CoverThomas}, a measure that has not been frequently used with continuous data. More importantly, we can provide a probabilistic interpretation of our measure's values. It captures the expected relative prediction loss incurred when predicting $Y$ while ignoring the value of $X$. Due to this interpretation as a measure of predictive ability, we named our measure PREDEP, an acronym standing for PREdictive DEPendence measure.

The PREDEP range is the $[0,1]$ interval. If PREDEP has a value of, for example, 0.5, this means that knowing $X$ can reduce the predictive loss of $Y$ by 50\%. The two extreme values in this $[0,1]$ range characterize extreme probabilistic behavior. PREDEP is zero if, and only if, $X$ and $Y$ are independent, whereas PREDEP equal to 1 means that $Y$ is perfectly predicted by $X$. Among other properties of PREDEP, in specific special cases such as the bivariate Gaussian distribution, we found a functional relationship between PREDEP, the correlation coefficient, and the mutual information measure.

Another contribution of our work is the development of a bootstrap method to estimate PREDEP with empirical data. Specifically, in the conditional distribution component of PREDEP, we need a method to handle the small number of data resulting from conditioning $X$ to a specific value. We can interpret the measure components as a certain convolution density function evaluated at zero, and we explore this interpretation to establish an estimator based on a bootstrap method.

To evaluate our measure empirically, we compute PREDEP across more than 90 thousand real datasets in addition to other synthetic datasets. PREDEP's performance was compared with several leading available alternatives (Pearson correlation coefficient \cite{pearson1895vii}, mutual information measure \cite{CoverThomas}, Maximal information criterion \cite{reshef2011detecting}, the distance correlation \cite{szekely2007measuring, szekely2009brownian}, the Hilbert-Schmidt Independence Criterion \cite{gretton2005measuring}, and Maximum Mean Discrepancy \cite{tolstikhin2016minimax}). This comparison demonstrated that our measure provides additional insights relative to these alternative association measures.

In summary, the main contributions of this paper are:
\begin{itemize}
\item PREDEP, a new association measure to evaluate the ability of $X$ to predict $Y$. The measure is not specialized for any specific type of association (such as linearity or monotonicity). It can capture arbitrarily complex non-linear relationships and it has a probabilistic interpretation as the percentage loss in prediction ability of $Y$ when we ignore $X$. 
\item Proofs of several theoretical properties of the measure.
\item A proposed method to estimate PREDEP with empirical data based on 
a bootstrap method coupled with a convolution density evaluation interpretation of the measure. 
\item An extensive empirical evaluation of the measure is based on more than 90 thousand real datasets and numerous synthetic datasets. We compared PREDEP with the six most relevant alternative association measures. 
\end{itemize}

\section{Related work}
\label{sec:relwork}

\begin{figure}[t!]
    \centering
        \includegraphics[width=\columnwidth]{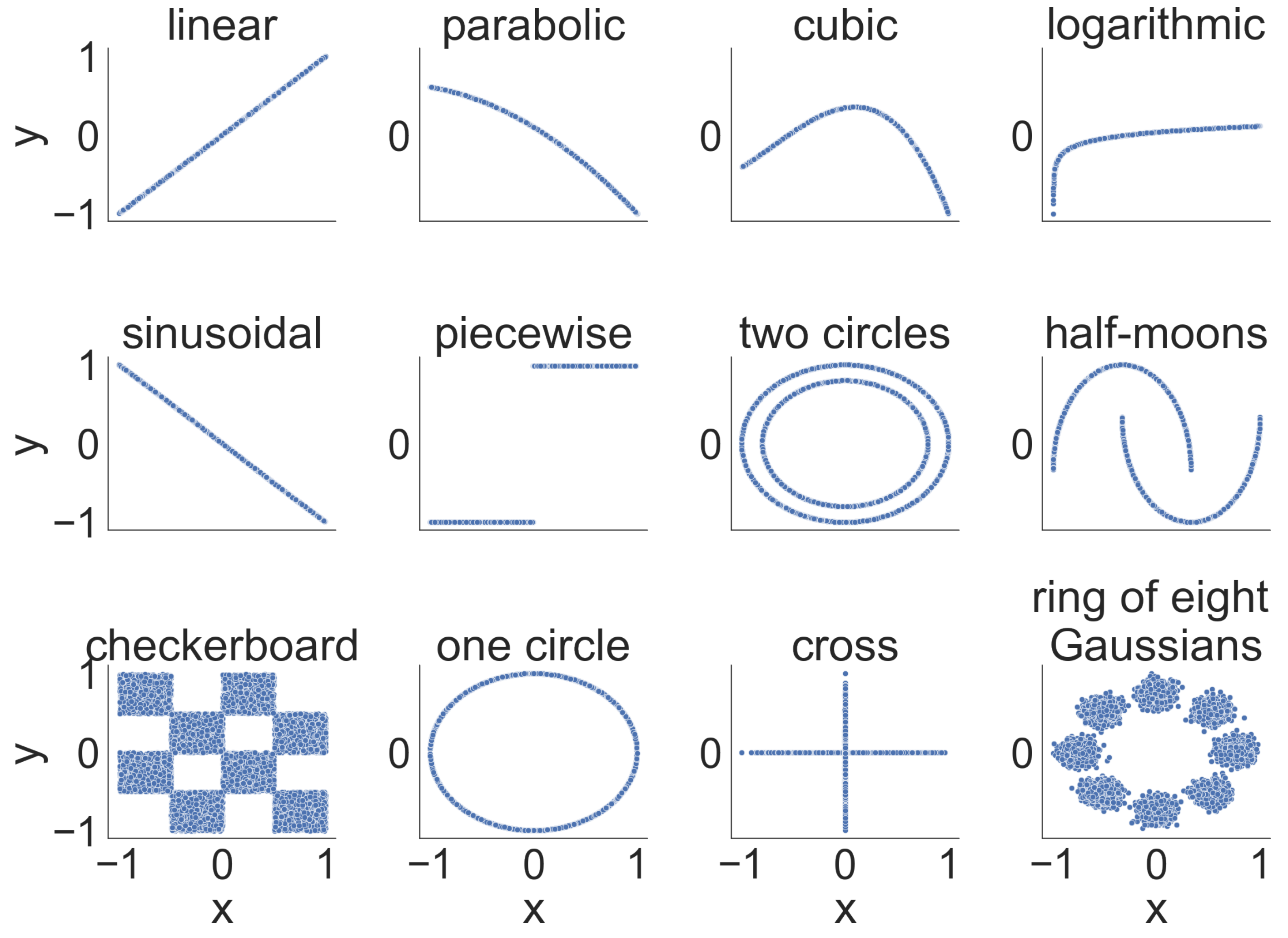}
    \caption{Visualization of functional and non-functional relationships used for benchmarking.}\vspace{-1em}
    \label{fig:vizu}
\end{figure}

Figure \ref{fig:vizu} shows examples of different types of association between random variables $X$ and $Y$. The first row of plots are cases in which the conditional distribution of $Y$ is of the form $(Y|X=x) = h(x) + \varepsilon$, where the function $h$ represents a hypothesis and $\varepsilon$ has an independent distribution of $X$, such as $N(0, \sigma^2)$. However, this simple functional relationship between $X$ and $Y$ is not the only way they can be dependent. In the second row of plots, we see examples where $X$ and $Y$ are associated but do not have a functional relationship akin to $(Y|X=x) \approx h(x) + \varepsilon$.

The Pearson correlation coefficient is one of the oldest and most popular measures of dependence between two random variables. While it captures well the linear association between $X$ and $Y$, it performs poorly in non-linear relationships. Spearman's rank correlation \cite{spearman1987proof} and Kendall's tau \cite{kendall1938new} were more flexible measures as they aimed to detect the presence of monotonic relationships but not necessarily linear relationships. Hoeffding's $D$ \cite{HoeffdingD} was the first attempt to test dependence between two random vectors without restricting their type of relationship but it has not shown great performance \cite{clark2013comparison}. 

Around the same time as \cite{HoeffdingD}, Mutual Information, $I(X, Y)$, enabled the measurement of non-functional dependencies in a general manner. However, estimating $I(X, Y)$ from data is quite challenging \cite{moon1995estimation,darbellay1999estimation,kraskov2004estimating}.

Another earlier attempt at measuring non-linear and non-monotonic dependence is the maximal correlation between $X$ and $Y$ proposed by \cite{gebelein1941statistische} and defined as 
$\text{mCor}(X,Y) = \sup_{f, g} \text{Cor}(f (X); g(Y))$ 
where the supremum is taken over all functions 
$f$, $g$. $\text{Cor}$ is Pearson's correlation. mCor is estimated through the Alternating Conditional Expectations algorithm \cite{breiman1985estimating}. Recently, \cite{nguyen2014multivariate} extended the maximal correlation coefficient to deal with random vectors. 

Promising proposals to measure dependence were made recently. The first is the Distance Correlation (\textit{dcor}) proposed by \cite{szekely2007measuring} and \cite{szekely2009brownian}. It has been used recently by the image-processing community 
for few-shot classification problems \cite{xie2022joint, xin2023hyperspectral, 
zhang2021local, li2023adaptive} and for object detection 
\cite{wu2023group, bilinski2014representing,bkak2016exploiting}. 
It has also become popular among the computer networks 
community to deal with website fingerprinting attacks \cite{zou2022efficient} and 
among the reliability community for machinery fault diagnosis \cite{yang2022novel}. 

%Cortando a explicação de dcor

%The distance correlation between $X$ and $Y$ is based on the distance between their joint characteristic function and the product of the marginal characteristic functions. One advantage of the distance correlation lies in its intricate theoretical foundation, which can be readily estimated using sample data. Given a set of $n$ pairs $(x_i, y_i),$ the approach involves constructing $n \times n$ matrices $A$ and $B,$ with the centered pairwise $x$'s and $y$'s distances. The estimation of the distance covariance involves computing the inner Frobenius product: $\text{dcov}^2(X, Y) = \langle \tilde{A} , \tilde{B} \rangle_{F} / n^2.$ Finally, the distance correlation is derived by taking the square root of the ratio $\text{dcor}^2(X, Y) = \frac{\text{dcov}^2(X, Y)}{\sqrt{\text{dcov}^2(X, X) \text{dcov}^2(Y, Y)}}$. \cite{bottcher2019distance} proposed an extension of the distance correlation measure that operates in a multivariate context.
The \textit{dcor} measure has gained significant popularity due to its performance in handling noisy functional relationships between $X$ and $Y$. However, it falls short of effectively capturing simple non-functional relationships. Consider, for instance, the scenario where $(X, Y)$ represents a random point selected from a narrow circle (see Figure \ref{fig:vizu}, plot $(2,4)$). We have $\text{dcor}^2(X, Y) \approx 0.2$ despite the strong functional association between $X$ and $Y$. Similarly, for the checkerboard pattern in Figure \ref{fig:vizu}, we have $\text{dcor}^2(X, Y) \approx 0.2$, yet it is evident that there exists an association between the variables.

Another recent and notably influential proposal is the Maximal Information Coefficient (MIC), introduced by \cite{reshef2011detecting}. The calculation of MIC involves an optimization process over a series of non-uniform grids. Each grid involves counting the points within the cells and then computing the mutual information as in a contingency table. The MIC algorithm exhaustively explores all available grids up to a certain maximal grid resolution, contingent upon the sample size. The resultant maximal mutual information value obtained from these grids is then normalized to a range between 0 and 1.

MIC is equitable: the measure yields similar scores for relationships of comparable noise levels but differing types. However, this finding has been controversial \cite{kinney2014equitability, ReshefAOAS2018, Reshef_StatSci2020}. \cite{kinney2014equitability} and \cite{liu2022detecting} suggest that the shortcomings of MIC might be attributed to using small sample sizes. Another issue with MIC has been its relatively low statistical power~\cite{simon2014comment}.

\section{A new association measure for continuous variables}
\label{sec:new-measure}

An association measure is a single number that summarizes the strength of the relationship between two random variables $X$ and $Y$. Additionally, it could indicate the ability of $X$ to predict $Y$. For \textit{categorical} random variables, the classical Goodman-Kruskal’s $\tau_{b}$ association measure was proposed by~\cite{goodman1954measures}. It calculates the reduction in the probability of incorrectly predicting the category of $Y$ when information is supplied about the category of $X$. It is remarkable for being one of the few dependence measures to have a clear empirical interpretation~\cite{reimherr2013quantifying}. This is the basis from which we depart to obtain PREDEP. The appendix presents a more detailed description and discussion of $\tau_{b}$. 

Consider a generic contingency table (see Figure \ref{fig:ContabAndGrid}) of two categorical variables, $Y$ (rows) and $X$ (columns), with cell counts equal to $n_{ij}$ and total $n_{\cdot \cdot} = \sum_{ij} n_{ij}$. The row and column totals are $n_{i\cdot} = \sum_j n_{ij}$ and $n_{\cdot j} = \sum_i n_{ij}$, respectively. Suppose that we condition on the row totals and that we randomly assign $n_{1\cdot}$ items to row 1 without replacement, $n_{2\cdot}$  to row 2 without replacement, etc. That is, we randomly shuffle the items in the rows keeping the row marginal distribution. 
After this random shuffle, some items will remain in their original row, while others will be assigned to different rows. We consider the latter situation as an assignment error.
The expected number $A$ of errors in these assignments is 
$A=n_{\cdot \cdot} \sum_i{p_{i \cdot}(1-p_{i \cdot})}$ where $p_{i\cdot} = n_{i\cdot}/n_{\cdot \cdot}$
(see the appendix for more details).  
For fixed $n_{\cdot \cdot}$, the value of $A$ is maximized when $Y$ has a uniform distribution.

Suppose now that the true columns label is known and the $n_{\cdot \cdot}$ items are partitioned into groups 
according to their columns. We carry out the same random row shuffling procedure within each column. Consider the $j$-th column, whose total is $n_{\cdot j}$. Condition on $n_{1j}, n_{2j}, \ldots$, the numbers in the rows of the $j$-th column, take the $n_{\cdot j}$ items and assign them randomly such that $n_{1j}$ items are assigned to row 1, $n_{2j}$ are assigned to row 2, etc. Writing $p_{ij} = n_{ij}/n_{\cdot \cdot}$, the expected number of errors by summing over the columns is $ B=\sum\limits_j{n_{\cdot j}\sum\limits_i{\frac{p_{ij}}{p_{\cdot j}}\left(1-\frac{p_{ij}}{p_{\cdot j}}\right)}}$ (see appendix for more details). The Goodman-Kruskal association measure $\tau_b$ is $(A-B)/A$. It measures the proportional reduction of the $Y$ class prediction error when the class of $X$ is known.  It is an asymmetrical measure, meaning $\tau_b$ typically differs if we reverse the conditioning by assigning the items to columns rather than rows (see appendix). Its value indicates the effectiveness of knowing one variable’s class in predicting the other variable’s class. In contrast to other measures, such as the chi-square statistic for contingency tables, $\tau_b$ is invariant to scale: if we increase the sample size by a multiple of $M$, the value of $\tau_b$  remains unchanged. In contrast, the chi-square statistic is multiplied by 
$M$,   even though all relative ratios between $n_{ij}$ remain constant.

To motivate the generalization of $\tau_b$ for continuous variables, we assume a sample of $N$ 
independent pairs $\{ (x_k, y_k), k=1, \ldots, N \}$ of realization of the random vector $(X,Y)$ with 
continuous joint probability density $f_{XY}(x,y)$ and marginal densities $f_{X}(x)$ and $f_{Y}(y)$. Visualize these points on a scatterplot overlayed by a regular grid with vertical and horizontal spacing between grid lines, denoted as $\Delta_y$ and $\Delta_x$, respectively (see Figure \ref{fig:ContabAndGrid}). 

To generate a contingency table, we count the number of sample points in each cell of a predefined grid. Although this table allows us to compute the Goodman and Kruskal $\tau_{b}$, it has a significant limitation: its dependence on the grid spacing. A similar challenge was encountered by \cite{reshef2011detecting} with their MIC measure, which they addressed by maximizing the mutual information over a collection of grids. In our approach, we not only avoid this dependence on grid spacing but also aim to define a theoretical measure that is expressed in terms of the underlying joint density, $f_{XY}(x,y)$, rather than relying on an empirical dataset. To address this dual requirement, we examine the asymptotic behavior of the Goodman-Kruskal measure as the grid spacing approaches zero and let the sample size $N \rightarrow \infty$.
We begin by deriving the analog of $A$ in the Goodman-Kruskal measure. However, instead of computing the number of errors when randomly shuffling points into vertical bins (equivalent to the rows in the contingency table), we focus on the expected \textit{rate of correct assignments}. This marks a fundamental departure from the classical $\tau_b$ measure.

Ignore initially the information on $X$, the variable represented on the horizontal axis. We wish to classify the $N$ points into the bins on the vertical axis according to their marginal distribution using a multinomial distribution, where the $i$-th vertical bin has probability $n_{\cdot i}/N$ with $n_{\cdot i}$ being the number of points whose $y$-coordinate lies within the $i$-th bin. Assume a small $\Delta_y$, large $N$, and the usual regularity conditions for the density $f_Y(y)$. If $\tilde{y}_i$ is the mid-center of the $i$-th vertical bin, we have $n_{i \cdot}/N \approx \mathbb{P}(Y \in \text{bin}_i) \approx f_Y(\tilde{y}_i) \Delta_y + o(\Delta_y^2)$.

Let $D$ represent the event that a random point $(x_k, y_k)$ is correctly assigned to its vertical bin and $C_i$ be the event that the $i$-th bin is the correct one. Of course, the $C_i$ events are disjoint and their union is the sample space. Hence, the rate $A$ of correct assignments in the classes (or bins) of length $\Delta_y$ in the vertical axis (per unit $y$-length) is given by
\begin{align*}
    A(\Delta_y) &= \frac{\mathbb{P}\left( D \right)}{\Delta_y} =  \frac{1}{\Delta_y} \sum_i \mathbb{P}\left( D \cap C_i \right) \\
     &= \frac{1}{\Delta_y} \sum_i  \mathbb{P}\left( D | C_i \right) \mathbb{P}\left( C_i \right) \: . 
\end{align*}

We have  $\mathbb{P}\left( D | C_i \right) = n_{i \cdot}/N \approx f_Y(\tilde{y}_i) \Delta_y + o(\Delta_y^2)$. Also, $\mathbb{P}( C_i ) \approx f_Y(\tilde{y}_i)\Delta_y + o(\Delta_y^2)$. Therefore, the rate of correct assignments, as $\Delta_y \rightarrow 0$ and $N \rightarrow \infty$, is 

\begin{equation}
  \begin{aligned}
    A(\Delta_y) &= \sum\limits_i f_Y^2(\tilde{y}_i)\Delta_y +o(\Delta_y^2) \\
       &\rightarrow \int f_Y^2(y)dy = \mathbb{E} \left[ f_Y(Y) \right] = S_Y 
  \end{aligned}
  \label{eq:SY}
\end{equation}

The measure $S_Y$ is essentially the $L^2$ norm of $f_Y$ in the space of real square-integrable functions. 
However, its prediction interpretation is the main motivation for its use: $S_Y$ is the probability \textit{rate} of successful prediction of $Y$ when one knows the marginal distribution $f_Y(y)$. Without any additional information from $X$, it is the expected value of the density $f_Y$ evaluated at a random value $Y$. The rate is calculated by a unit of $y$-length. For example, if $Y$ is income measured in dollars, $S_Y$ is calculating the probability of successful prediction of $Y$ \textit{per} dollar. 

We shall now evaluate how this prediction of $Y$ can be improved when there is additional information on $X$. The events $D$ and $C_i$ refer now to a modified assignment mechanism, one that is defined on bins in the support of $X$. Since $X$ is a continuous random variable, we discretize its support with bins of length $\Delta_x$. 
We repeat the calculation of the rate $A(\Delta_y)$ in each discrete $x$-bin, sum over all these $x$-bins and take the limit with $\Delta_y, \Delta_x \rightarrow 0$ and $N \rightarrow \infty$. More specifically, 
let $E_j$ be the event that the randomly selected point belongs to the $j$-th bin in the support of $X$. The midpoint of this $j$-th bin is $\tilde{y}_j$. We calculate $A(\Delta_y| X \in I_j)$ the same measure $A(\Delta_y)$ that was defined previously, but now we condition on the event $\{ X \in I_j \}$. We average the results to obtain 
\begin{small}
\begin{align*}
B(\Delta_y, \Delta_x) &= \frac{1}{\Delta_y} \sum_j \mathbb{P}\left( D | X \in I_j \right) \mathbb{P}\left( X \in I_j \right) \\
                      &= \frac{1}{\Delta_y} \sum_j \sum_i \mathbb{P}\left( D \cap C_i | X \in I_j \right) \mathbb{P}\left( X \in I_j \right) \\
                      &= \frac{1}{\Delta_y} \sum_j  \mathbb{P}\left( X \in I_j \right) \sum_i \mathbb{P}\left( D | C_i \cap \{X \in I_j \} \right)\\
                      & ~~~~~~~~~ \cdot \mathbb{P}\left( C_i | X \in I_j \right) 
\end{align*}
\end{small}
Using again the usual approximations, we can write 
\begin{small}
    \begin{align}
        B(\Delta_y, \Delta_x) &=  \sum_j f_X(\tilde{x}_j) \Delta_x \sum_i f^2_{Y|X \in I_j}(\tilde{y}_i| X\in I_j) \Delta_y  \nonumber \\
                              &+  \frac{o(\Delta_y^2) o(\Delta_x)}{\Delta_y} \nonumber \\
                              &\rightarrow \int \left[ \int f^2_{Y|X}(y|X=x) ~ dy \right] f_X(x) ~ dx  \nonumber \\
                              &= E_X \left[ E_{Y|X} f(Y|X) \right] = S_{Y|X}
    \label{eq:SYX}            
    \end{align}    
\end{small}
as $\Delta_x \rightarrow 0$, $\Delta_y \rightarrow 0$ and $N \rightarrow \infty$. 
The PREDEP association measure we propose is the relative additional predictive ability of using the $X$ variable to predict $Y$:
\begin{equation}
   \alpha_{Y\mid X} = (S_{Y|X} - S_Y)/S_{Y|X} \: .
   \label{eq:alpha}
\end{equation}

%É um pouco desnecessario OK
%Thus, PREDEP measures the expected relative \textit{prediction loss} of $X$ one incurs by ignoring the $Y$ distribution. 
%We have an analogous score when we swap variables: 
%\begin{equation}\label{eq:alpha2}
%\alpha_{X\mid Y} = \frac{S_{X|Y} - S_X}{S_{X|Y}}
%\end{equation}

We have an analogous score when we swap variables. As Goodman and Kruskal's $\tau_b$ association measure, PREDEP is not symmetric and $\alpha_{X\mid Y}$ is the expected relative prediction loss of $X$ one incurs by ignoring the $Y$. As discussed in the next section, being asymmetric is an advantage of our score. For simplicity, we shall use $\alpha$ when discussing our properties below as they are valid in both directions.

\section{Properties of $\alpha$}
\label{sec:properties}

\noindent {\bf P1:} $\alpha \in [0,1]$. The interpretation of $\alpha$ as a relative loss is granted only if $\alpha \in [0,1]$. This is indeed true. The proof is in the supplementary material.

\noindent {\bf P2:} $\alpha =0$ if, and only if, $X$ and $Y$ are independent. The proof is in the supplementary material.

\noindent {\bf P3:} Connection with R\'{e}nyi information measure. The quadratic R\'{e}nyi entropy is given by $H_{2}(f) = - \log \left( \int f^{2}(y) dy \right) $. Therefore, $S_Y = e^{-H_2(f_Y)}$. We are not aware of the R\'{e}nyi entropy being used to measure dependence for continuous variables as we have done in this paper.

\noindent {\bf P4:} PREDEP is asymmetric. However, this characteristic is not a drawback. Consider $(y|x) = x^2 + \epsilon$, where $\epsilon \sim \mathcal{N}(0, \sigma^2)$ with $\sigma \approx 0$ and $x \sim \mathcal{U}(-1,1)$. Having knowledge of $x$ naturally leads to a fairly accurate prediction of $y$ with a minimal margin of error. On the other hand, the reverse isn't true. This function is not bijective, meaning that a single value of $y$ can be generated by at least two (or potentially more due to the presence of noise) different values of $x$. The PREDEP score $\alpha_{Y|X}$ is $0.913$, while $\alpha_{X|Y}$ is $0.696$. It is easier to predict $Y$ from $X$ than $X$ from $Y$. The PREDEP score accurately reflects this asymmetry in predictability.

\noindent \textbf{Some analytical case} 
Consider the bivariate normal distribution where $X\sim N(\mu_X,\sigma^2_X)$, $Y \sim N(\mu_Y,\sigma^2_Y)$, and the correlation coefficient is $\rho$. In the supplementary material, we show that $\alpha_{Y|X} = \alpha_{X|Y} =\alpha$ and $\alpha = 1 - \sqrt{1 - \rho^2}$, establishing a direct link between $\alpha$ and $\rho$. In this Gaussian case, PREDEP exhibits symmetry irrespective of the sequence of conditioning. Additionally, mutual information is intertwined with $\alpha$, as $I(X,Y)= -\frac{1}{2} \log(1 - \rho^2) = -\log(1 - \alpha)$.

\section{Learning $\alpha$ from data}

%In practice, $\alpha$ must be estimated.
One simple way to estimate $S_Y$ in Equation (\ref{eq:SY}) is to substitute $f_Y$ by a smooth kernel density estimate $\hat{f}_{Y}$ and then integrate its square value numerically over the region:
$ \hat{S}_Y = \int \hat{f}_{Y}^2(y) dy$. 
However, this method proves insufficient when we attempt to estimate $S_{Y|X}$ in Equation (\ref{eq:SYX}) due to the necessity of estimating $f(Y| X=x)$ for every $x$ within a fine grid. Unless we are working with a very large dataset, there may not be enough data points within the narrow interval $[x-\Delta_x, x+\Delta_x]$ to accurately estimate the conditional density of $(Y | X=x)$.

To tackle this issue, we utilize a bootstrap procedure by leveraging the fact that $E\left[ f_Y(Y) \right]$ represents a convolution density at the value zero. Suppose that $Y_1$ and $Y_2$ are independent copies of the continuous random variable $Y$. The density of 
$W=Y_1 - Y_2$ is $ f_W(w) = \int f_Y(y-w) f_Y(y) dy $
which, evaluated at $w=0$, gives $f_W(0) = \int f_Y^2(y) dy = E\left[ f_Y(y) \right]$. Therefore, if we have a large sample of the variable $W=Y_1-Y_2$ and use this large sample to calculate a kernel-based estimate $\hat{f}_W(w)$ of its density $f_W(w)$, the value $\hat{f}_W(0)$ will be an estimate of $E\left[ f_Y(y) \right]$.

Suppose we have a sample $Y_1, \ldots, Y_n$ of $Y$. Generate a large bootstrap sample with $b$ elements given by $(Y_{11}, Y_{21}), (Y_{12}, Y_{22}), \ldots, (Y_{1b}, Y_{2b})$ where all $Y_{ij}$ are selected independently and with equal probability from $Y_1, \ldots, Y_n$. Define $W_i \equiv Y_{1i} - Y_{2i}$. Based on these $b$ values of the random
variable $W_i$, estimate the density $f_W(w)$ using a kernel-density estimate and evaluate it at $w=0$. 

We repeat this procedure for $S_{Y|X}$ considering the subsample of values in bins along the $x$-axis. We employ a Hierarchical Clustering with $k = \sqrt{N}$ clusters to determine the bins. While not theoretically rigorous, this rule offers a practical balance between simplicity and detail. The algorithm for PREDEP is in Appendix~\ref{alg:pseudo}. To address uncertainty, we propose Bootstrap Confidence Intervals (CI). PREDEP is an expected value. Thus, a Bootstrap is an adequate choice for a non-parametric CI. 

\section{Empirical evaluation}
\label{sec:empiricalwork}

The comprehensive comparison undertaken in \cite{reshef2011detecting} significantly influenced our selection of measures for our empirical assessment. We focused on the best-performing measures in their study: MIC and the distance correlation (dcor). We also present comparisons involving the traditional Pearson and mutual information measures despite their lackluster performance in \cite{reshef2011detecting}, as well as the Conditional Mutual Information (CMI)~\cite{molavipour2021neural} and Hilbert-Schmidt Independence Criterion (HSIC)~\cite{gretton2005measuring}. We also considered Maximum Mean Discrepancy (MMD)~\cite{tolstikhin2016minimax}. However, it is inappropriate for our problem as it verifies if marginal distributions are similar. 
%For our assessment of dcor, we employed the implementation provided by the \texttt{dcor} Python library \cite{ramos2023dcor}. In the case of MIC, we utilized the \texttt{minepy} Python library \cite{albanese2013minerva}.

\noindent \textbf{Synthetic data:} We follow \cite{reshef2011detecting} to generate our synthetic data:  $(X, Y^*)$ with $X \sim U(-1,1)$, $Y=h(X)$ where $h$ is a function and $Y^* = Y + \epsilon$ with $\epsilon \sim U(-\delta, \delta)$. The amount of noise is measured by $1-r^2 = 1-\text{cor}^2(Y, Y^*)$ and the functions selected are shown in the first two rows of plots in Figure 
\ref{fig:vizu}. They are: Linear ($Y = \theta_0 X + \theta_1$); Parabolic ($Y = \theta_2 X^2 + \theta_1 X + \theta_0$); Cubic ($Y = \theta_3 X^3 + \theta_2 X^2 + \theta_1 X + \theta_0$); Logarithmic ($Y = \theta_0 \ln(X + 1)$); Sinusoidal ($Y = \theta_1 \sin(\theta_0 X)$); Piecewise ($Y = \min(\max(1/X, \theta_0), \theta_1)$). In all cases, the coefficients $\theta_i$ are drawn from a uniform distribution spanning (-1, -1), except in piecewise case, where $\theta_0 \sim U(-3,0)$ and $\theta_1 \sim U(0,3)$. We consider different values for $\delta$ to see how the metrics respond to the introduction of incremental noise.  We used 1000 noise levels for each scenario. For each noise level, we generated a dataset with 1000 $(x,y)$ points.

 \begin{figure}[t!]
    \centering
        \includegraphics[width=\columnwidth]{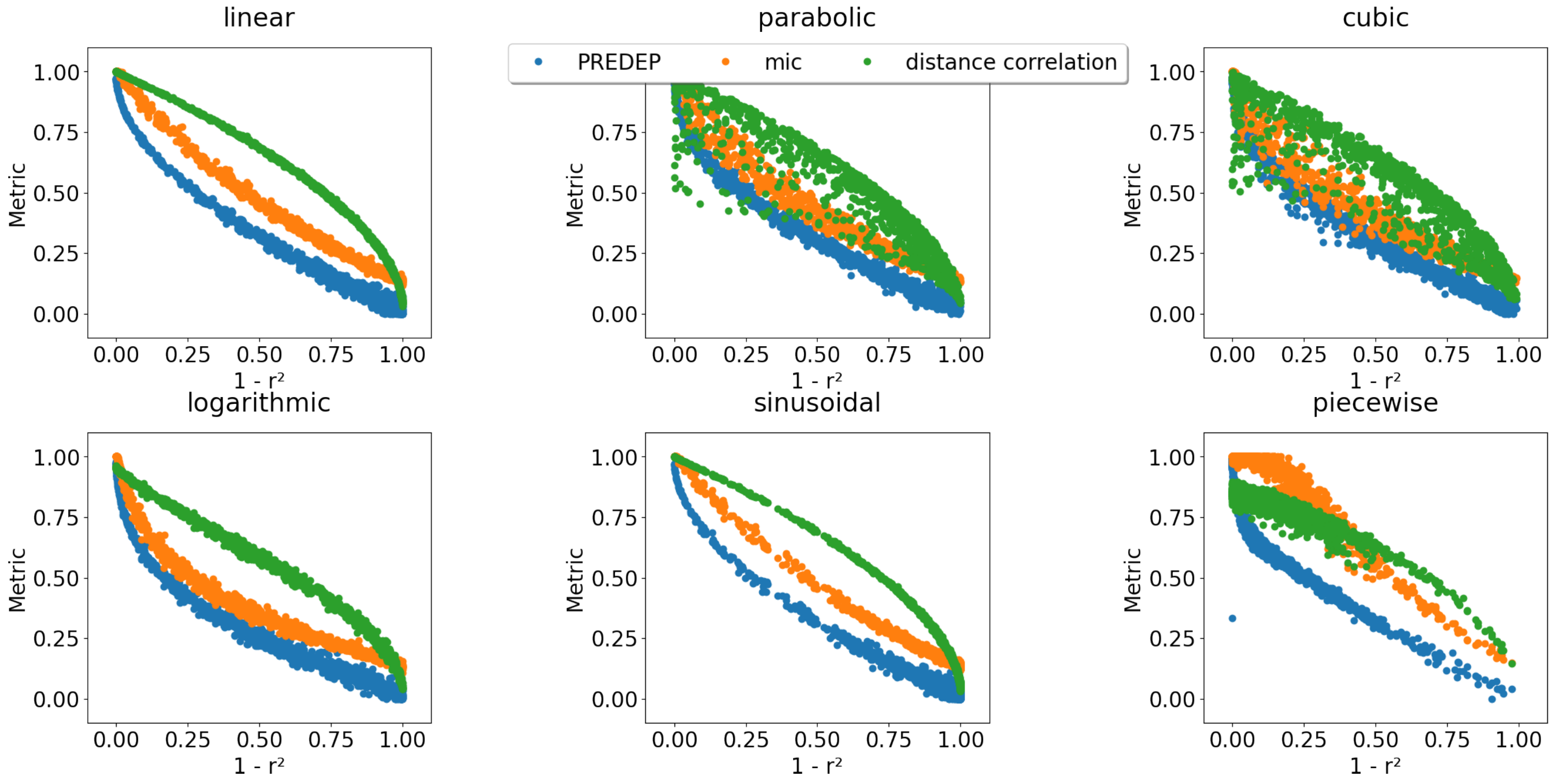}
    \caption{Behavior of MIC and PREDEP in a functional relationship.}
    \label{fig:functional_relationship}
\end{figure}

Table \ref{tab: w_noise} shows the average values of the measures with the noiseless data with the Pearson correlation coefficient($|r|$),  Spearman coefficient ($|r_s|$), maximal correlation ($\rho_{max}$), distance correlation ($\rho_{dist}$), information coefficient of correlation ($r_1$), maximal information coefficient (MIC) and PREDEP (PD).

The data, originally from \cite{Rainio2022}, had MIC recalculated and PREDEP calculated for analysis. Both metrics show high values in functional relationships, indicating similar behaviors. In piecewise relationships, the metrics start to diverge, likely due to discontinuities. The greatest differences appear in non-functional relationships. PREDEP has a higher value in the cross-shaped relationship, whereas MIC is higher in circular and checkerboard patterns. These differences highlight the importance of using both metrics for a comprehensive understanding, as they measure different aspects of the data. The measures HSIC and MIC lack a clear interpretation beyond stating that higher values indicate stronger dependence. However, these values are often unintuitive and difficult to interpret—a drawback that PREDEP aims to address.

\begin{table}[t!]
 \centering \caption{\footnotesize The average values of the coefficients with noiseless data. }
 %Linear (Lin); Logarithmic (Log); Cubic (Cub); Quadratic (Qua); Sinusoidal (Sin); Piecewise (Piw); Cross-Shaped (Cro); Circular (Cir); Checkerboard (Che).}
 
\resizebox{\columnwidth}{!}{
\begin{tabular}{llllllllll}
\toprule
       Model & $|r|$ & $|r_s|$ &$\rho_{m}$&$\rho_{d}$& $r_1$ & MIC & PD & CMI & HSIC \\
\midrule
      Lin & 1.00 & 1.00 & 1.00 &  1.00 & 0.99 & 1.00 & 0.97 & 5.68 & 123.13\\
 Log & 0.99 & 1.00 & 1.00 &  0.96 & 0.99 & 1.00 & 0.98 & 5.30 & 0.97\\
       Cub & 0.78 & 1.00 & 0.99 &  0.84 & 0.99 & 1.00 & 0.98 & 4.27 & 100.97\\
   Qua & 0.06 & 0.03 & 1.00 &  0.86 & 0.97 & 1.00 & 0.98 & 4.92 & 105.12\\
  Sin & 0.05 & 0.12 & 0.98 &  1.00 & 0.92 & 1.00 & 0.97 & 4.70 & 129.19\\
   Piw & 0.44 & 0.50 & 0.98 &  0.86 & 0.97 & 1.00 & 0.88 & 2.46 & 95.30\\
Cro & 0.00 & 0.00 & 0.93 &  0.30 & 0.80 & 0.44 & 0.86 & 0.46 & 18.67\\
    Cir & 0.03 & 0.03 & 0.99 &  0.16 & 0.96 & 0.67 & 0.31 & 1.89 & 14.81\\
Che & 0.00 & 0.00 & - &  0.20 & - & 0.46 & 0.22 & 0.65 & 2.57\\
\bottomrule
\end{tabular}
}
\label{tab: w_noise}
\end{table}

\begin{comment}
\begin{table}[t!]
 \centering \caption{\footnotesize The average values of the coefficients with noiseless data.}
 \resizebox{8cm}{!}{
\begin{tabular}{llllllll}
\toprule
       Model & $|r|$ & $|r_s|$ &$\rho_{max}$&$\rho_{dist}$& $r_1$ & MIC & PD \\
\midrule
      Linear & 1.000 & 1.000 & 1.000 &  1.000 & 0.994 & 1.000 & 0.970 \\
 Logarithmic & 0.987 & 1.000 & 1.000 &  0.955 & 0.994 & 1.000 & 0.976 \\
       Cubic & 0.779 & 1.000 & 0.995 &  0.843 & 0.994 & 1.000 & 0.980 \\
   Quadratic & 0.056 & 0.033 & 1.000 &  0.856 & 0.969 & 1.000 & 0.982 \\
  Sinusoidal & 0.056 & 0.123 & 0.984 &  1.000 & 0.919 & 1.000 & 0.971 \\
   Piecewise & 0.441 & 0.504 & 0.979 &  0.856 & 0.973 & 1.000 & 0.884 \\
Cross-shaped & 0.001 & 0.001 & 0.931 &  0.301 & 0.800 & 0.441 & 0.864 \\
    Circular & 0.027 & 0.032 & 0.995 &  0.162 & 0.958 & 0.666 & 0.307 \\
Checkerboard & 0.000 & 0.000 & -     &  0.203 & -     & 0.462 & 0.223 \\
\bottomrule
\end{tabular}
}
\label{tab: w_noise}
\end{table}
\end{comment}

Figure \ref{fig:functional_relationship} illustrates the behavior of PREDEP, MIC, and dcor as the noise $1-r^2$ increases. Across all examined functional relationships, the behavior of PREDEP was similar. Variations lie in the speed at which the metric tends towards zero and the degree of metric deviation. Here, PREDEP is more similar to MIC than to distance correlation. Concerning MIC, the main difference lies in the standard deviation. At low noise levels, PREDEP exhibits a small deviation, which increases as the noise grows. MIC behaves oppositely.  

There doesn't seem to be a direct ordering relationship between the metrics. However, a pattern emerges where distance correlation values are larger than MIC, which in turn has values larger than PREDEP. It is important to remark that the metrics are measuring 
\textit{different} aspects of the data and this implies different scales and values, as it will be clear in the non-functional cases.  

\begin{figure}[t]
    \centering
        \includegraphics[width=\columnwidth]{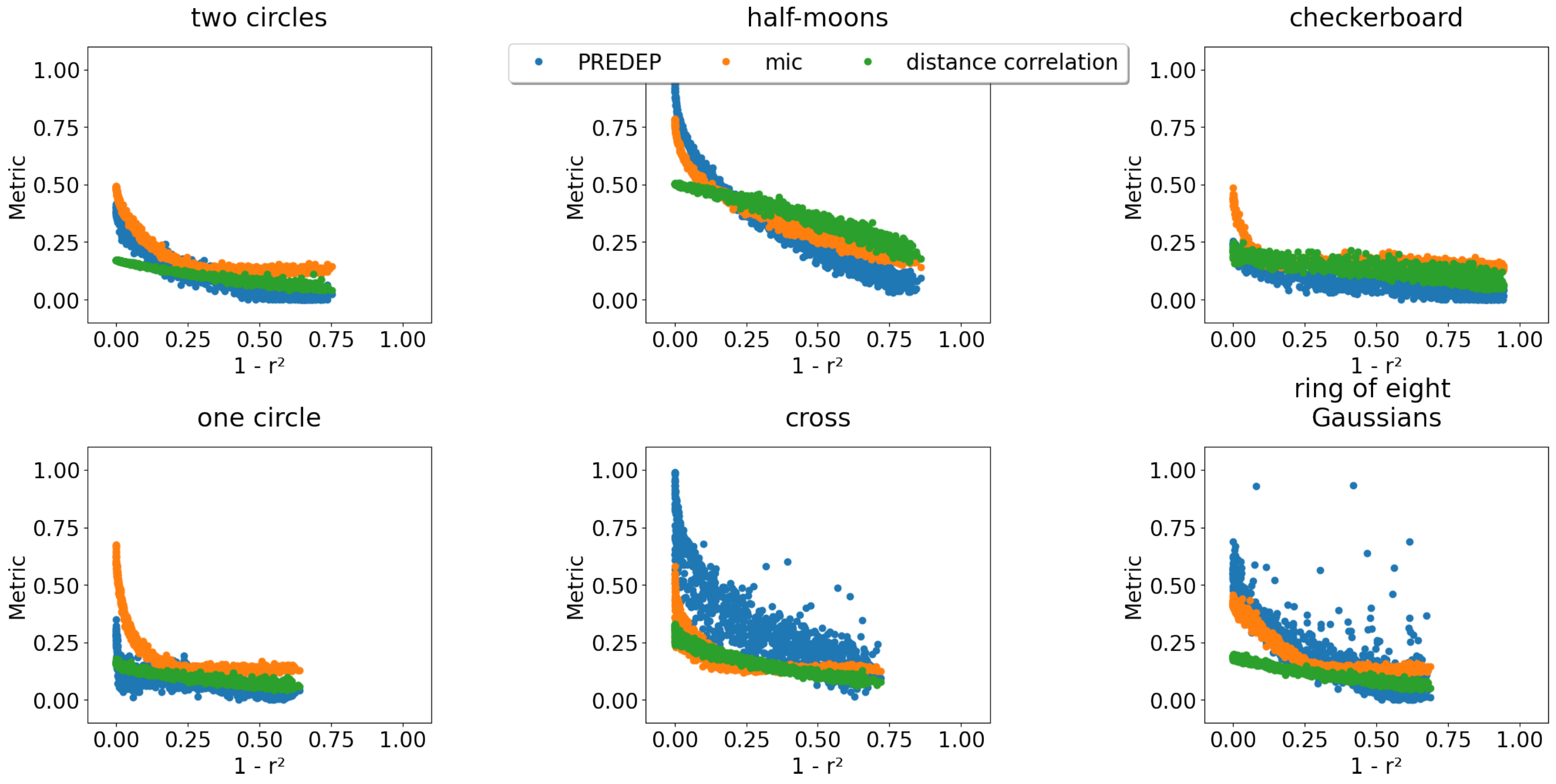}
    \caption{Behavior of MIC and PREDEP in non-functional relationships.}
    \label{fig:non_functional_relationship}
\end{figure}

\noindent \textbf{Non-functional relationship.}
The last two rows of plots of Figure \ref{fig:vizu} show the 
type of non-functional relationships we used. Noise is added to $Y$ in the form of $U(-\delta, \delta)$. 
Figure \ref{fig:non_functional_relationship} illustrates the metrics' behavior for non-functional relationships, highlighting significant differences. PREDEP shows the highest value in the cross pattern and the lowest in the checkerboard pattern due to its measurement of predictive ability. In the cross pattern, knowing $Y$ (if not near zero) lets us predict $X \approx 0$, except when $Y$ is close to zero. In the checkerboard pattern, knowing $Y$ halves the interval of possible $X$ values, but significant uncertainty remains.
MIC and dcor indicate the presence of an association between $X$ and $Y$  but do not indicate that the degree of this association is low in terms of predictability capacity. These metrics measure deviations in the joint distribution from independence, while PREDEP measures predictive capability. Consider the PREDEP increase when passing from one circle to the two circles case. In the smaller circle's radius, knowing $X$ offers little gain, but between the larger and smaller radii, it narrows down possible values from four to two.

The supplementary material analyzes a more significant number of metrics in the noiseless case. We also compared MIC, dcor, and PREDEP for copula models.

\noindent \textbf{Real datsets} %\label{sec:empiricalwork}
In this section, we utilized the dataset obtained from the World Health Organization (WHO), which was previously employed \cite{reshef2011detecting}. It comprises 357 social, economic, political, and health indicators collected from a comprehensive set of 202 countries in 2009. As it does not contain complete data for all indicators across all countries, we applied our analysis exclusively to pairs of indicators that had values recorded for at least 50 countries. This selection process resulted in a total of 96,980 ordered pairs of variables. Although the number of pairs of variables is large, the number of points in each scatterplot is relatively low compared to those used in the synthetic dataset analysis. 

\begin{figure}[t!]
    \centering
    \subfloat[][PREDEP vs MIC]{\raisebox{0.12cm}{\includegraphics[width=0.22\textwidth]{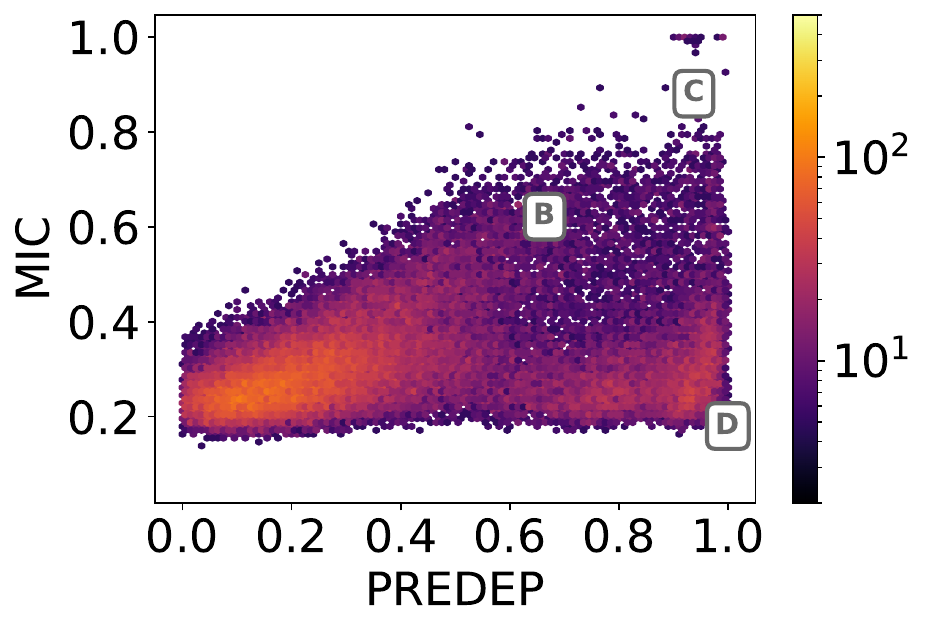}}}\quad
    \subfloat[][Mid MIC/PREDEP]{\includegraphics[width=0.22\textwidth]{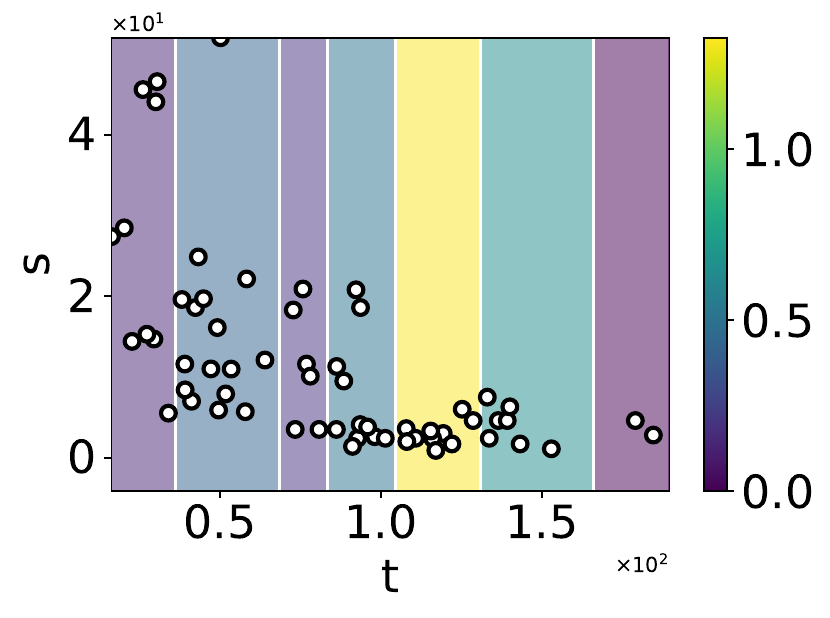}}\quad
    \subfloat[][High PREDEP/MIC]{\includegraphics[width=0.22\textwidth]{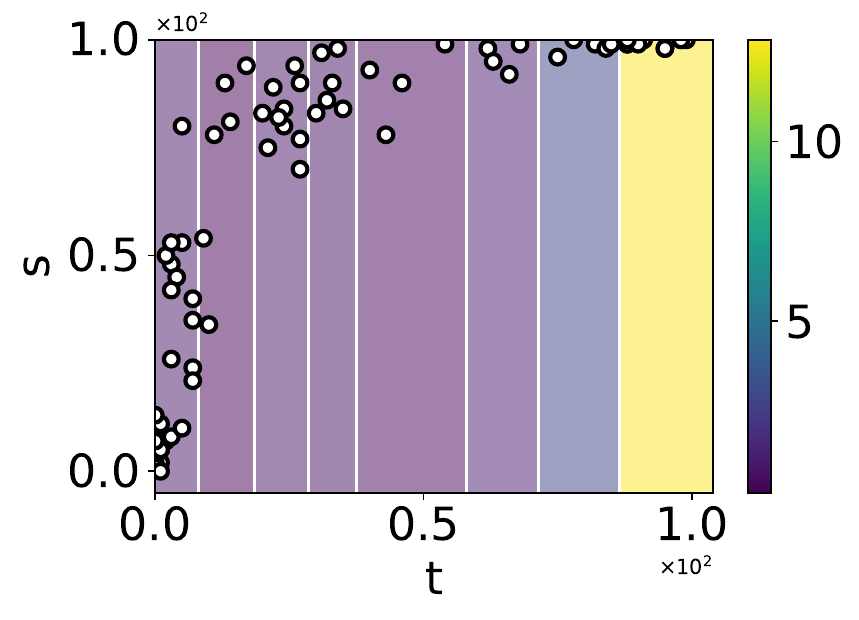}}\quad
    \subfloat[][Low M. vs High P. ]{\includegraphics[width=0.22\textwidth]{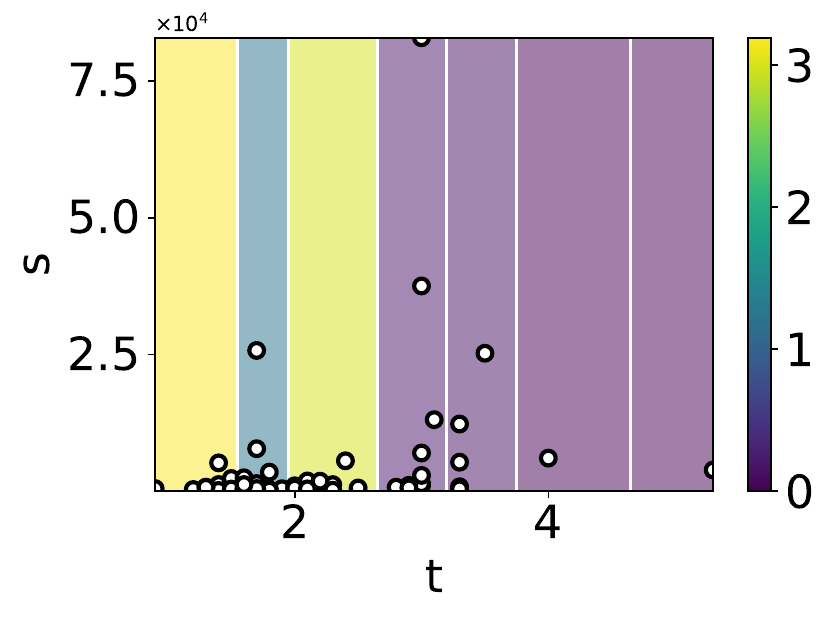}}
    \caption{Application of PREDEP and MIC metrics to the 96,980 selected indicator pairs dataset.}
    \label{fig:who_data}
\end{figure}

We calculated PREDEP, MIC, DCOR, HSIC, CMI and the Pearson correlation coefficient to the variable pairs to comprehend how these metrics perform on the same real-world dataset. We started presenting the MIC and PREDEP results in Figure \ref{fig:who_data}.

When observing Figure \ref{fig:who_data}A, we can perceive the presence of two concentrations of data points. The first concentration, steeper and with points that share similar values in both coordinates, demonstrates a relative agreement between the two measures for a substantial portion of pairs. The second concentration, in the lower right corner of the graph, reveals that for another subset of variable pairs, the PREDEP values obtained are higher than the MIC values.

To better understand the results of the comparison between the two metrics, we further examined pairs of variables in different regions of the graph. The distributions of values for three such pairs are presented in Figures \ref{fig:who_data}B, \ref{fig:who_data}C, and \ref{fig:who_data}D. The colored intervals show the binning used in our implementation (a hierarchical clustering on $X$). The background colors of the scatterplots represent the value of  $S_{Y|X=x}/ S_Y$, where $x$ is the bin, according to the scales located to the right of each plot. When $S_{Y|X=x}/S_Y$ increases, it means that we have a high predictive power on the bin.

The variable pairs used in each of these B, C, and D thumbnail plots are described in the Appendix. The first example, in Figure \ref{fig:who_data}B, illustrates a case of moderate association agreement by both methods, with a PREDEP of 0.66 and MIC of 0.62. Here, we can observe an actual relationship where higher values of $X$ suggest lower values of $Y$, albeit with some dispersion.
In Figure \ref{fig:who_data}C, there's also an agreement between the two metrics, but the association is very strong. This aligns with intuition when observing the scatterplot, indicating a pronounced increasing trend between the variables, reminiscent of a logarithmic curve. Lastly, Figure \ref{fig:who_data}D portrays an interesting example: the pair of variables received a PREDEP value of 0.86, while MIC yielded only 0.28. The dispersion of points shows that the $Y$ values are concentrated in a narrow lower range, with a few outliers. Moreover, in the lighter intervals, it's noticeable that the points are even more tightly clustered along the \(y\)-axis, justifying the high PREDEP value.

\begin{figure}[t!]
    \centering
    \subfloat[][PREDEP vs DC]{\includegraphics[width=0.22\textwidth]{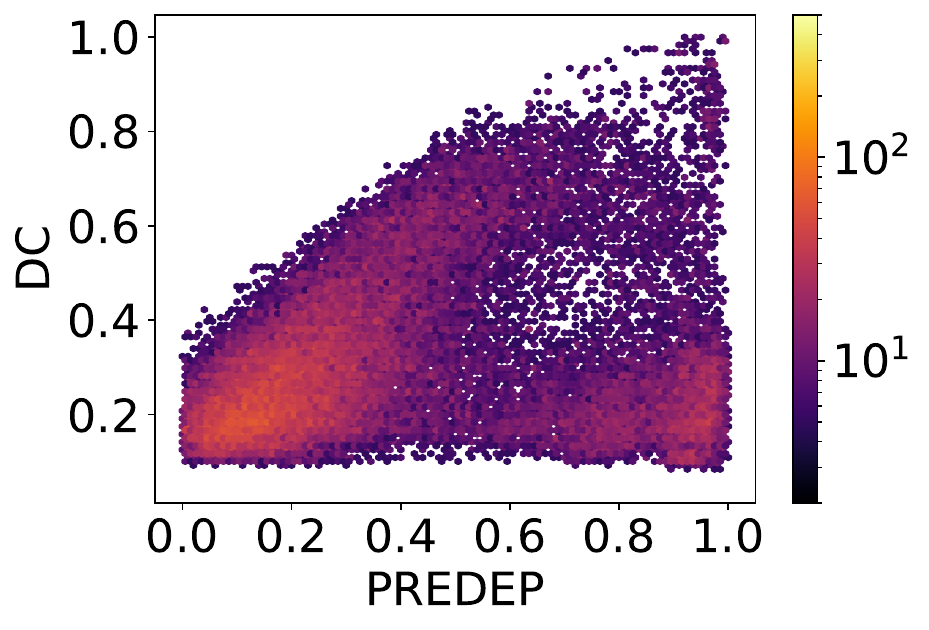}} \quad
    \subfloat[][PREDEP vs HSIC]{\includegraphics[width=0.22\textwidth]{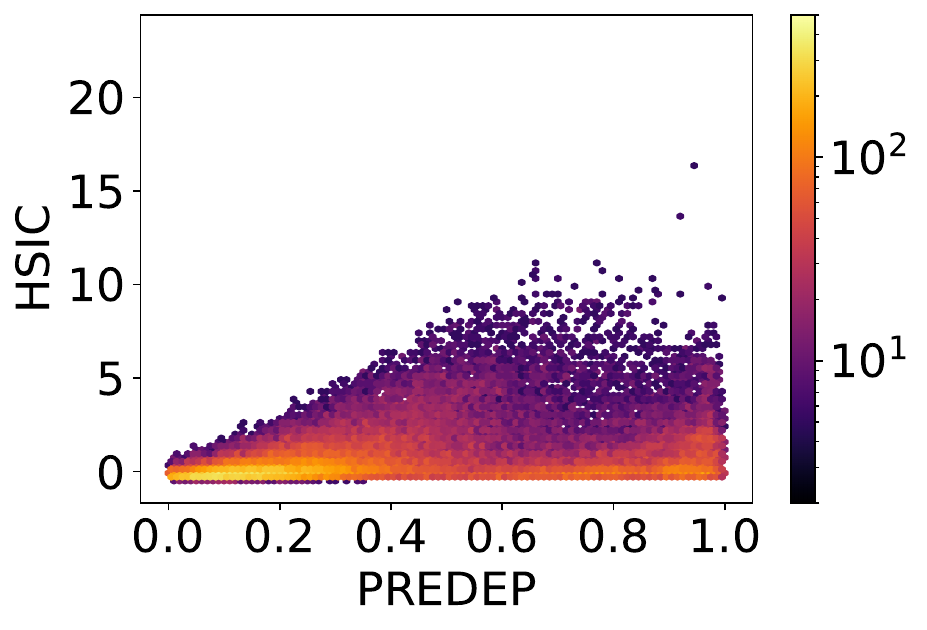}} \quad
    \subfloat[][PREDEP vs CMI]{\includegraphics[width=0.22\textwidth]{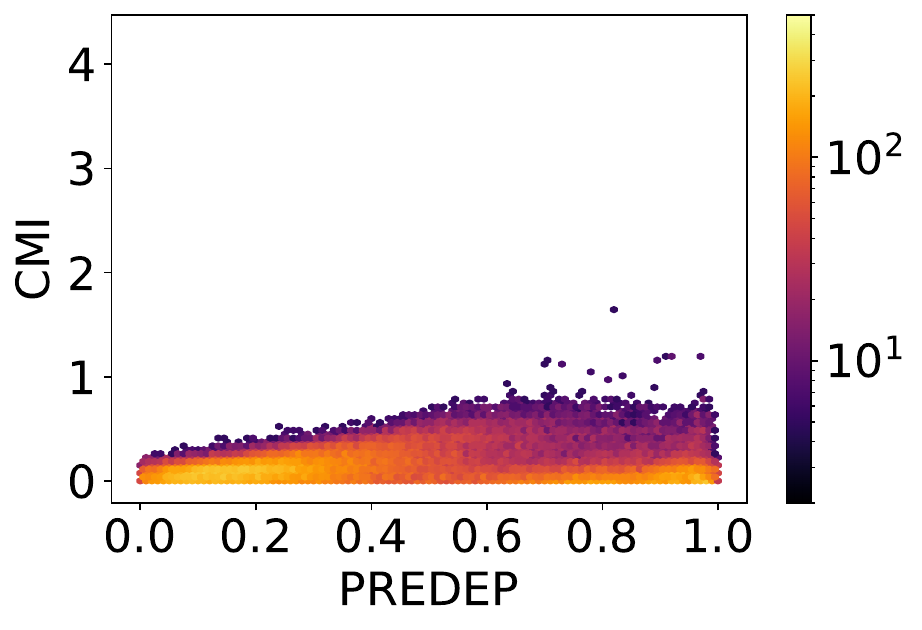}} 
    \caption{Comparison of PREDEP with DC, HSIC, and CMI on the WHO dataset}
    \label{fig:who_data_3}
\end{figure}

Another approach used with this dataset was applying Distance Correlation to the selected variable pairs, compared to PREDEP in Figure \ref{fig:who_data_3}A. The distribution of data points reveals two distinct clusters, similar to Figure \ref{fig:who_data}A. The larger, steeper cluster indicates agreement between the two metrics, while the lower-right cluster shows pairs with high PREDEP values but low Distance Correlation values.
Figure \ref{fig:who_data_3} also compares PREDEP with Conditional Mutual Information (CMI) and the Hilbert-Schmidt Independence Criterion (HSIC). Their behavior is similar to that of MIC in Figure \ref{fig:who_data}, with the main difference being the scale of the Y-axis. An extensive analysis comparing PREDEP and MIC with the Pearson correlation coefficient is in section E of the Appendix.

\section{Conclusions}

Our empirical evaluation shows that MIC, PREDEP, and dcor tend to agree but have different scales. The reason is that they capture different aspects of the association between $X$ and $Y$. The other two measures focus on measuring the distance 
between the joint probability distribution to the distribution obtained by the product of the marginal distributions. Our proposed PREDEP measures the expected relative prediction \textit{loss} one incurs by ignoring the distribution of $Y$ when predicting $X$. This predictive dependence measure can capture 
arbitrarily complex non-linear relationships. However, the more relevant property of $\alpha$ is its interpretation in terms of how much we improve 
on the prediction of $X$ given that we know the value of $Y$. 
We proved a series of properties of PREDEP and proposed a bootstrap method for its estimation. Although we have focused on bivariate distributions, in principle it is simple to extend the definition to measure dependence between two random vectors. However, equipped with accurate density and conditional density estimates for multidimensional data (e.g., pairs of variables where $\mathbf{X} \in \mathbb{R}^k$ and $\mathbf{Y} \in \mathbb{R}^j$), the PREDEP score remains the same. That is, $S_X = \mathbb{E} \left[ f_X(X) \right]$ and $S_{X|Y} = E_Y \left[ E_{X|Y} f_{X|Y}(X|Y) \right]$ remain unchanged. What does change are the density estimators $f(\cdot)$. In these more challenging cases, we may  estimate PREDEP using estimators like Mixture Density Networks, Kernel Mixture Networks, or Normalizing Flows.

As with any association measure, our measure should be seen as a measure of causal association. Nevertheless, PREDEP stands out from other traditional measures like MIC and dcor due to its non-symmetric nature. By quantifying the asymmetry between variables, PREDEP allows us to identify which variable, say $X$, is more predictive of $Y$ than vice versa. This feature may be useful in exploratory analysis aimed at discovering potential causal relationships.

In our view, we should abandon the search for a single 
best metric, and embrace the idea that more than one association measure has to be used to capture the nuances of complex modern datasets. PREDEP complements existing tools and should be welcomed as a good addition to our toolbox of association measures. Its unique predictive interpretation makes it a valuable addition to our set of association measures.

%\begin{contributions} % will be removed in pdf for initial submission 
					  % (without ‘accepted’ option in \documentclass)
                      % so you can already fill it to test with the
                      % ‘accepted’ class option
%    Briefly list author contributions. 
%    This is a nice way of making clear who did what and to give proper credit.
%    This section is optional.
%
%    H.~Q.~Bovik conceived the idea and wrote the paper.
%    Coauthor One created the code.
%    Coauthor Two created the figures.
%\end{contributions}

%\begin{acknowledgements} % will be removed in pdf for initial submission,
%						 % (without ‘accepted’ option in \documentclass)
%                         % so you can already fill it to test with the
%                         % ‘accepted’ class option
%    Briefly acknowledge people and organizations here.
%
%    \emph{All} acknowledgements go in this section.
%\end{acknowledgements}

% References
% Any choice of citation style is acceptable as long as you are consistent. 
% It is permissible to reduce the font size to \verb+small+ (9 point)
% when listing the references.
% Note that the Reference section does not count towards the page limit of 9 pages.
\bibliographystyle{unsrt}
\bibliography{completename}

\newpage 

\appendix

\begin{center}
    \Large{\textbf{Appendix}}
\end{center}

\section{Goodman-Kruskall association measure for categorical variables}

In a series of papers \cite{goodman1954measures, goodman1959measures, goodman1963measures, goodman1972measures}, Goodman and Kruskal introduced several measures of association between categorical random variables. Most of these measures have a probabilistic interpretation giving them an intuitive meaning. Goodman-Kruskal's coefficient (denoted as $\tau_{b}$) is the most popular. Consider a random sample of $n$ items classified according to 
two categorical variables $Y$ and $X$ with $r$ and $c$ categories, respectively. The general form of a contingency table is exemplified in Table \ref{tab:GeneralContingency}. The observed frequency or count in category $i$ of the row variable and category $j$ of the column variable is denoted by $n_{ij}$. The total number of observations in category $i$ of the row variable is $n_{i\cdot} = \sum_j n_{ij}$, and the total number of observations in category $j$ of the column variable is $n_{\cdot j} = \sum_i n_{ij}$. These are referred to as marginal totals. 
$\tau_{b}$ is obtained using the principle of proportional reduction in error in the prediction of the row category when we know the column category.  

\begin{table}[h]
 \centering \caption{General form of a contingency table }
\label{tab:GeneralContingency}
\begin{tabular}{ccccccc|c}\\ \hline
\hline
\multicolumn{ 3}{c}{Rows}&\multicolumn{ 4}{c}{Columns (Variable $X$)}&\\ \cline{4-8}
\multicolumn{ 3}{c}{(Variable $Y$) }&1 &  2 &  $\cdots$ &  c &Total\\ \hline
&\multicolumn{ 2}{c}{1}&$n_{11}$    &   $n_{12}$&   $\cdots$        &$n_{1c}$&$n_{1.}$      \\
&\multicolumn{ 2}{c}{2}&$n_{21}$&   $n_{22}$&   $\cdots$        &   $n_{2c}$&$n_{2.}$\\
&\multicolumn{ 2}{c}{$\vdots$}&$\vdots$ &$\vdots$&$\ddots$& $\vdots$&$\vdots$\\
&\multicolumn{ 2}{c}{r}&$n_{r1}$    &   $n_{r2}$    &   $\cdots$    &   $n_{rc}$&$n_{r.}$       \\ \hline
&\multicolumn{2}{c}{Total}&$n_{.1}$&$n_{.2}$&$\cdots$&$n_{.c}$&$ n_{\cdot \cdot}$ \\ \hline
\hline
\end{tabular}
\end{table}

Suppose we are asked to place the $n$ items in one of the $r$ row categories in such a manner that we end up with exactly $n_{1.}$ cases in row 1, $n_{2.}$ in row 2, up to $n{r.}$ in the $r$-th row. That is, we assign items randomly to the rows keeping the same row marginal counts as observed in the table. In this step, we do not care about the column label. It is clear that when we attempt to reallocate the items to their respective rows, we make errors. Items belonging to a certain row may end up being allocated to a different row.  
We can calculate the expected number $A$ of errors we expect to make in assigning the items to their correct row categories. For example, the probability of correctly assigning an item to row 2 is the proportion $p_{2.} = n_{2.}/n_{\cdot \cdot}$ of items that truly belong to row 2. Therefore, the expected total number of correct assignments to this row 2 is $n_{2.} p_{2.}$. We assign $n_{2.} = n_{\cdot \cdot} p_{2.}$ items to row 2. Therefore, the expected total number of errors is $n_{2.} (1 - p_{2.}) = n_{\cdot \cdot} p_{2.} (1 - p_{2.})$ in row 2. Summing over all the rows gives us the expected total number $A$ of errors: 
\[
 A=n \sum\limits_{i=1}^{r} {p_{i.}(1-p_{i.})} \: . 
 \]

Suppose now that we are informed which column each item belongs to, but we still do not know its correct row. We are also informed of the frequencies $n_{ij}$ in each column. Hence, the $n_{\cdot \cdot}$  items are partitioned into $c$ groups according to their column labels. We must attempt to allocate the $n_{\cdot j}$ elements of column $j$ to their correct rows, knowing that the row distribution is given by $n_{1j}, \ldots, n_{rj}$. Again, we will make errors, and the expected number of these errors, summing over the $c$ columns, is given by 
\[
 B=\sum\limits_j{n_{\cdot j}\sum\limits_i{\frac{p_{ij}}{p_{j.}}\left(1-\frac{p_{ij}}{p_{j.}}\right)}},
 \]
where $p_{.j} = n{_.j}/n_{\cdot \cdot}$ and $p_{ij} = n_{ij}/n_{\cdot \cdot}$.

Goodman-Kruskal's $\tau_{b}$ coefficient is given by 
\[ \tau_{b} = (A-B)/A \: . 
\] 
It can be proven that $\tau_{b}$ is in $[0,1]$. This coefficient measures the relative decrease in the probability of misclassifying the row variable when the column variable is known. For example, if $(\tau_{b} = 0.8$, it indicates an 80\% reduction in the probability of misclassifying the row category when information about the column category is provided. 

Obviously, we can also calculate $\tau_{b}$ exchanging rows and columns roles. This will measure the relative decrease in the probability of misclassifying the columns variable when the row variable is known. The two measures usually are not equal. Table \ref{tab:notsymmetricGK} is a small contingency table where the Goodman-Kruskal $\tau_b$ association measure is not symmetric, with one direction equal to 1 (indicating a perfect association), and the other direction having a substantially smaller association. Calculating $\tau_b$ for predicting rows given columns, we have $\tau_{b} = 1$ while the other direction gives $\tau_b=0.5$. This is not a drawback of the measure, quite the opposite. This example shows clearly that the conditional prediction of one variable given another is not a symmetric property and a prediction measure must reflect that. 

\begin{table}[h!]
\centering
\resizebox{\columnwidth}{!}{
\begin{tabular}{|c|c|c|c|c|c|}
\hline
      & Column 1 & Column 2 & Column 3 & Column 4 & Column 5 \\ \hline
Row 1 & 0        & 0        & 2        & 0        & 0        \\ \hline
Row 2 & 0        & 2        & 0        & 2        & 0        \\ \hline
Row 3 & 2        & 0        & 0        & 0        & 2        \\ \hline
\end{tabular}}
\caption{Contingency Table showing that Goodman-Kruskal association measure $\tau_b$ is not symmetric: a perfect association in one direction but only a moderate association in the other direction.}
\label{tab:notsymmetricGK}
\end{table}

\section{Computing Predep from Data} \label{alg:pseudo}

Our code is at: \url{https://anonymous.4open.science/r/predep-C0CE/}. The code used in our experiments is \url{https://anonymous.4open.science/r/predep-C0CE/predep_paper.py}. Another file, \url{https://anonymous.4open.science/r/predep-C0CE/predep.py}, is an implementation closer to the pseudo-code.

\begin{algorithm}[t!]
\scriptsize
\caption{Pseudocode for PREDEP $\alpha_{Y\mid X}$. Invert the parameters for $\alpha_{Y\mid X}$.}
\begin{algorithmic}[1]
\Function{predep}{$\mathbf{X}:$ array of data, $\mathbf{Y}$: array of data, $n_b = \lceil n \log(n) \rceil$: number of bootstrap samples}
    \State $\mathbf{Y}_1 \gets \text{bootstrap\_sample}(\mathbf{Y}, n_b)$
    \State $\mathbf{Y}_2 \gets \text{bootstrap\_sample}(\mathbf{Y}, n_b)$
    \State $\mathbf{W} \gets \mathbf{Y}_1 - \mathbf{Y}_2$
    \State $S_Y \gets \text{KDE}(\mathbf{W}).\text{pdf}(0)$
    \vspace{.75em}
    \State \Comment{$\mathbf{E}$ is an array of bin edges; the first index is where the first bin starts; the second one is where the first bin ends and the second bin starts; and so forth. This may be computed in equal spaces, log spaces, or via clustering on $\mathbf{X}$. Indexes start at zero in the loop below. We define bins by a Hierarchical Clustering on $\mathbf{X}$ using $k = \sqrt{n_y}$ clusters.}
    \State $\mathbf{E} \gets \text{compute\_bin\_edges}(\mathbf{X})$ 
    \vspace{.75em}
    \State $S_{Y|X} \gets 0$
    \For{$i \gets 1$ to $\text{length}(\mathbf{E}) - 1$}
        \State $bg \gets \mathbf{E}[i - 1]$
        \State $ed \gets \mathbf{E}[i]$
        \vspace{.75em}
        \State $\mathbf{Y}_{\mid \mathbf{X}} \gets 
        \text{filter}(\mathbf{X}, \mathbf{Y}, bg, ed)$ \Comment{Filters the values from $\mathbf{Y}$ on the indexes of $\mathbf{X}$ that are inside the $[bg, ed)$ range.}
        \State $\mathbf{Y}_{1\mid \mathbf{X}} \gets \text{bootstrap\_sample}(\mathbf{Y}_{\mid \mathbf{Y}}, n_b)$
        \State $\mathbf{X}_{2\mid \mathbf{X}} \gets \text{bootstrap\_sample}(\mathbf{Y}_{\mid \mathbf{X}}, n_b)$
        \State $\mathbf{W}_{\mathbf{Y}\mid\mathbf{X}} \gets \mathbf{Y}_{1\mid \mathbf{X}} - \mathbf{Y}_{2\mid \mathbf{X}}$
        \vspace{.75em}
        \State \Comment{$\text{length}(\mathbf{Y}_{\mid \mathbf{X}})$ counts the number of elements in the $[bg, ed)$ range. $p_y$ is thus the fraction of elements within the bin. This value estimates $P_Y(Y = y)$ and is aggregated below to empirically compute $E_Y \left[ E_{X|Y} f_{X|Y}(X|Y) \right]$.}
        \State $p \gets \frac{\text{length}(\mathbf{Y}_{\mid \mathbf{X}})}{\text{length}(\mathbf{Y})}$ 
        \vspace{.75em}
        %\State $p\_x\_mid\_y \gets \text{KDE}(DX\_mid\_Y).pdf(0)$
        \State $S_{Y|X} \gets S_{Y|X} + \big (p \cdot  \text{KDE}(\mathbf{W}_{\mathbf{Y}\mid\mathbf{X}}).\text{pdf}(0) \big)$

    \EndFor
    \vspace{.75em}
    \State $\alpha \gets \frac{S_{Y|X} - S_Y}{S_{Y|X}}$
    \State \Return $\alpha$
\EndFunction
\end{algorithmic}
\end{algorithm}

\section{Properties of $\alpha$: proofs and details}

\noindent {\bf P1:} $\alpha \in [0,1]$. 

\noindent \textbf{Proof:} Since $S_Y > 0$, 
we only need to prove that $S_{Y|X} \geq S_Y$.
We have
\[\int f^2_Y(y)  dy - \int\left(f_Y(y)-f_{Y|X}(y|x) \right)^2 dy\leq \int f_Y^2(y)  dy \]
where the equality holds only when $f_Y(y) = f_{Y|X}(y|x)$, almost surely in $y$ and $x$.
Expanding the square of the second term on the left-hand side, the first integral is canceled out, and the inequality
reduces to
\[ 2\int f_{Y|X}(y|x)f_Y(y)dy - \int f_{Y|X}^2(y|x)dy \leq \int f_Y^2(y) dy \]
Taking the integral with respect to $x$ on both sides of this last expression, we have:
\begin{small}
\begin{align*}
    \iint & \left[ 2 f_{Y|X}(y|x) f_Y(y) - f_{Y|X}^2(y|x) \right] f_X(x) dy dx  
    \\ & \leq \iint f_Y^2(y) f_X(x) dy dx \: .
\end{align*}
\end{small}
This implies that 
\begin{small}
\begin{align*}
   2 \int f_Y(y) & \int f_{X,Y}(x,y)dx dy-\iint f_{Y|X}^2(y|x) f_X(x)dy dx  
   \\ & \leq \int f_Y^2(y) dy \: ,  
\end{align*}
\end{small}
or 
\begin{small}
\begin{align*}
   2 \int f_Y(y)f_Y(y) dy & -\iint f_{Y|X}^2(y|x) f_X(x)dy dx  
   \\ & \leq \int f_Y^2(y) dy \: . 
\end{align*}
\end{small}
Therefore, we have
\begin{small}
\[ S_{Y|X} = \iint f^2_{Y|X}(y|x) f_X(x)dy dx \geq \int f_Y^2(y) dy = S_Y \: , \]
\end{small}
and since $S_Y > 0$, we see that $\alpha = 1 - S_Y/S_{Y|X}  \in [0,1]$.

\noindent {\bf P2:} $\alpha =0$ if, and only if, $X$ and $Y$ are independent. 

\noindent \textbf{Proof:} It is clear that, if $X$ and $Y$ are independent,  we have $f_{Y|X}(y|x) = f_Y(y)$ and obviously $S_{Y|X} = S_Y$. Therefore, $\alpha=0$. 

For the converse, assume that $\alpha=0$. Then,
\begin{small}
\begin{align*}
    0 &= S_{Y|X} - S_Y  \\
    & = \int \left[ \int f^2_{Y|X}(y|x) - f^2_{Y}(y)  \right] f_{X}(x) dx \\
    &= \iint f^2_{Y|X}(y|x)  f_{X}(x) dy dx - \iint f^2_{Y}(y)  f_{X}(x) dx \\
    &=  \iint \frac{f^2_{XY}(x,y)}{f_{X}(x)} dy dx
    - \int f^2_{Y}(y) dy \int f_{X}(x) dy \\
    &=  \iint \frac{f^2_{XY}(x,y)}{f_{X}(x)} dy dx
    - \int f^2_{Y}(y) dy \\
     &=  \iint \frac{f^2_{XY}(x,y)}{f_{X}(x)} dy dx
    - 2 \int f^2_{Y}(y) dy + \int f^2_{Y}(y) dy \\
    &=  \iint \frac{f^2_{XY}(x,y)}{f_{X}(x)} dy dx
    - 2 \int f_{Y}(y) \left[ \int f_{XY}(x,y) dx \right] dy \\ 
    & ~~~~~~~~~~  + \int f^2_{Y}(y) \left[ \int f_X(x) dx \right]dy \\
    &=  \iint \frac{f^2_{XY}(x,y)}{f_{X}(x)} dy dx
    - 2 \iint f_{Y}(y) f_{XY}(x,y) dy dx \\
    & ~~~~~~~~~~  + \iint f^2_{Y}(y) f_X(x) dy dx \\        
\end{align*}
\end{small}
\begin{small}
\begin{align*}
%    &=  \iint \left[ \frac{f^2_{XY}(x,y)}{f_{X}(x)}  
%    - 2 \frac{f_{Y}(y) f_{XY}(x,y) f_X(x)}{f_X(x)} + \frac{f^2_{Y}(y) f_X^2(x)}{f_X(x)} \right] dy dx \\ 
    &=  \iint \frac{f^2_{XY}(x,y) - 2 f_{Y}(y) f_{XY}(x,y) f_X(x) + f^2_{Y}(y) f_X^2(x)}{f_X(x)}  dy dx \\
    &= \iint \frac{\left[ f_{XY}(x,y) - f_Y(y) f_X(x) \right]^2} {f_X(x)}  dy dx \\ 
\end{align*}
\end{small}
As the integrated is a non-negative function, we have $\left[ f_{XY}(x,y) - f_Y(y) f_X(x) \right]^2 = 0$ almost surely. This implies that $f_{XY}(x,y) = f_Y(y) f_X(x)$ almost surely and we can conclude that $X$ and $Y$ are independent. 

\noindent {\bf P3:} Connection with R\'{e}nyi information measure. 
\textbf{Details:} 
For $a > 0$ and $a \neq 1$, the R\'{e}nyi entropy generalizes the Shannon entropy and it is defined as 
\[ H_{a}(f) = \frac{1}{1-a} \log \left( \int f^{a}(x) dx \right) \]
\cite{renyi1961measures, CoverThomas}. The case $a=2$ is called quadratic R\'{e}nyi entropy and it is equal to 
\[ H_{2}(f) = - \log \left( \int f^{2}(x) dx \right) \: . \]
Therefore, $S_X = e^{-H_2(f)}$. Unlike the discrete case, R\'{e}nyi entropy can be negative for continuous random variables, and so R\'{e}nyi entropy is typically only used for discrete variables. For example, let $X$ be the random variable defined on $[1, \infty)$ with density
$f(x)=3x^{-4}$. We have $H_2(f)=-0.2513$. This somewhat limits its interpretation, in contrast with our measure $\alpha$.

\noindent \textbf{Some analytical cases} 

\begin{figure}[h]
    \centering
        \includegraphics[scale=0.20]{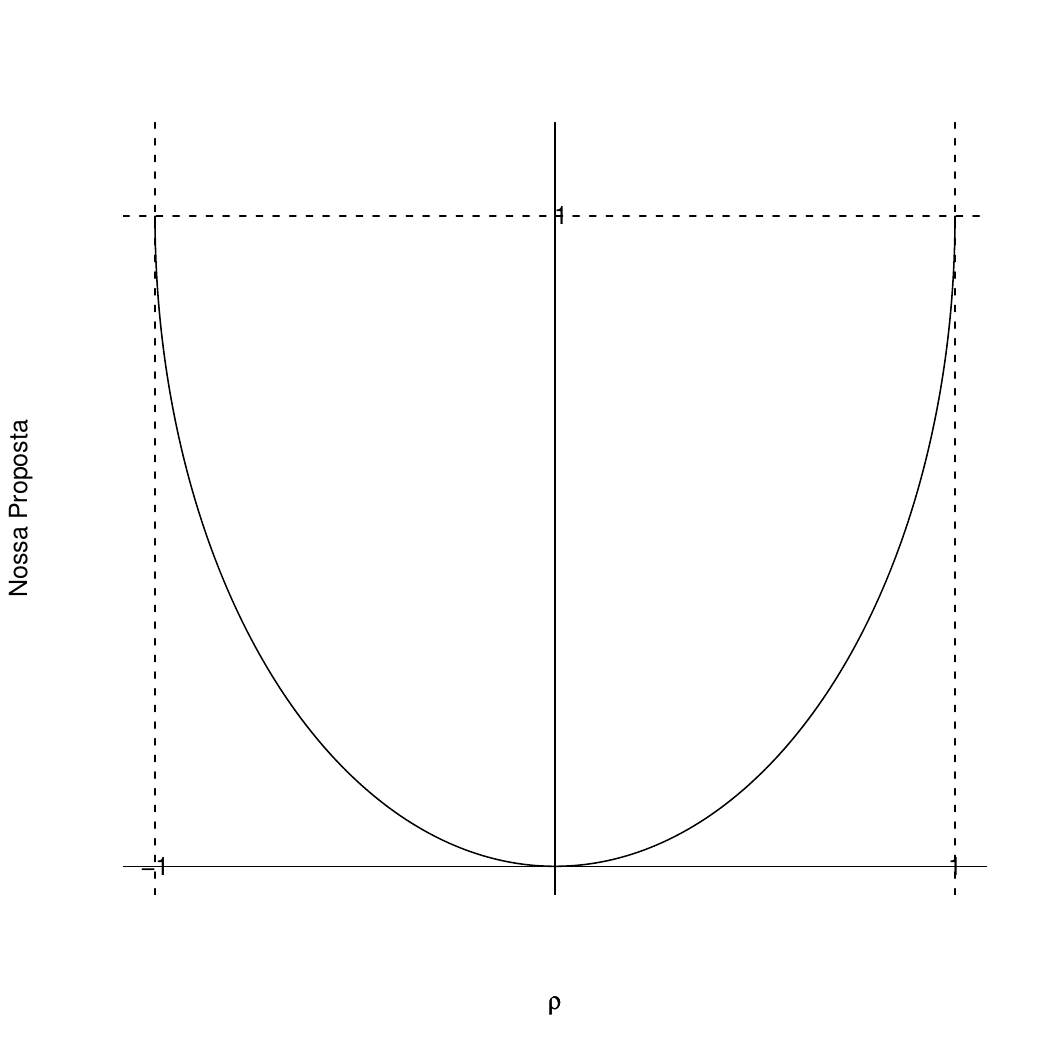}
        \includegraphics[scale=0.25]{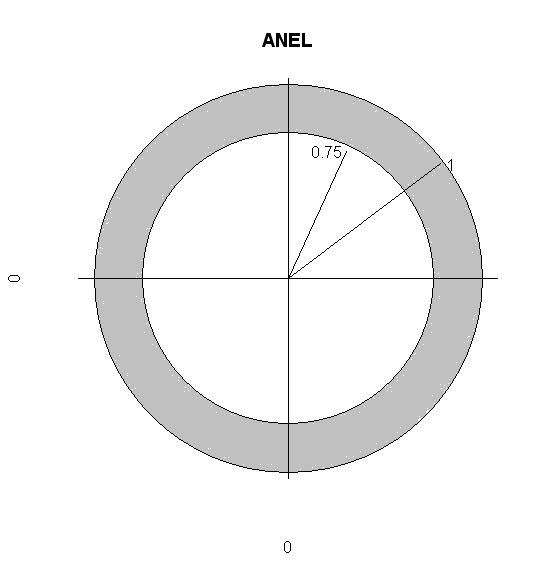}
    \caption{Left: Relationship between $\alpha$ and $\rho$ in the bivariate normal case. Right: 
    The vector $(X,Y)$ is uniformly distributed in the annulus. The Pearson correlation coefficient is zero but our PREDEP measure is greater than zero.}
    \label{fig:normalidade}
\end{figure}

Consider the bivariate normal distribution where $X\sim N(\mu_X,\sigma^2_X)$, $Y \sim N(\mu_Y,\sigma^2_Y)$, and the correlation coefficient $\rho$. 
It is a simple calculation to obtain
\[ S_Y = \int f^2_Y(y) dy = \frac{1}{2\sigma_Y\sqrt{\pi}} \]
and, since $Y|X=x$ is also a normal distribution, to obtain
\[ S_{Y|X} = \int \frac{1}{2\sigma_Y\sqrt{\pi (1-\rho^2)}} f_X(x) dx = \frac{1}{2\sigma_Y\sqrt{\pi (1-\rho^2)}} \]
Therefore,
\[ \alpha = 1 - \sqrt{1 - \rho^2} \: , \]
providing a simple relationship between $\alpha$ and $\rho$ (see left-hand side of Figure \ref{fig:normalidade}).
The right-hand side of Figure \ref{fig:normalidade} shows our second analytical example,
where $(X,Y)$ is uniformly distributed in a ring formed by the 
annulus between the circles $x^2 + y^2 = 3/4$ and $x^2 + y^2 = 1$. 

\section{Performance with copula models}

 \begin{figure}[h]
    \centering
        \includegraphics[width=\columnwidth]{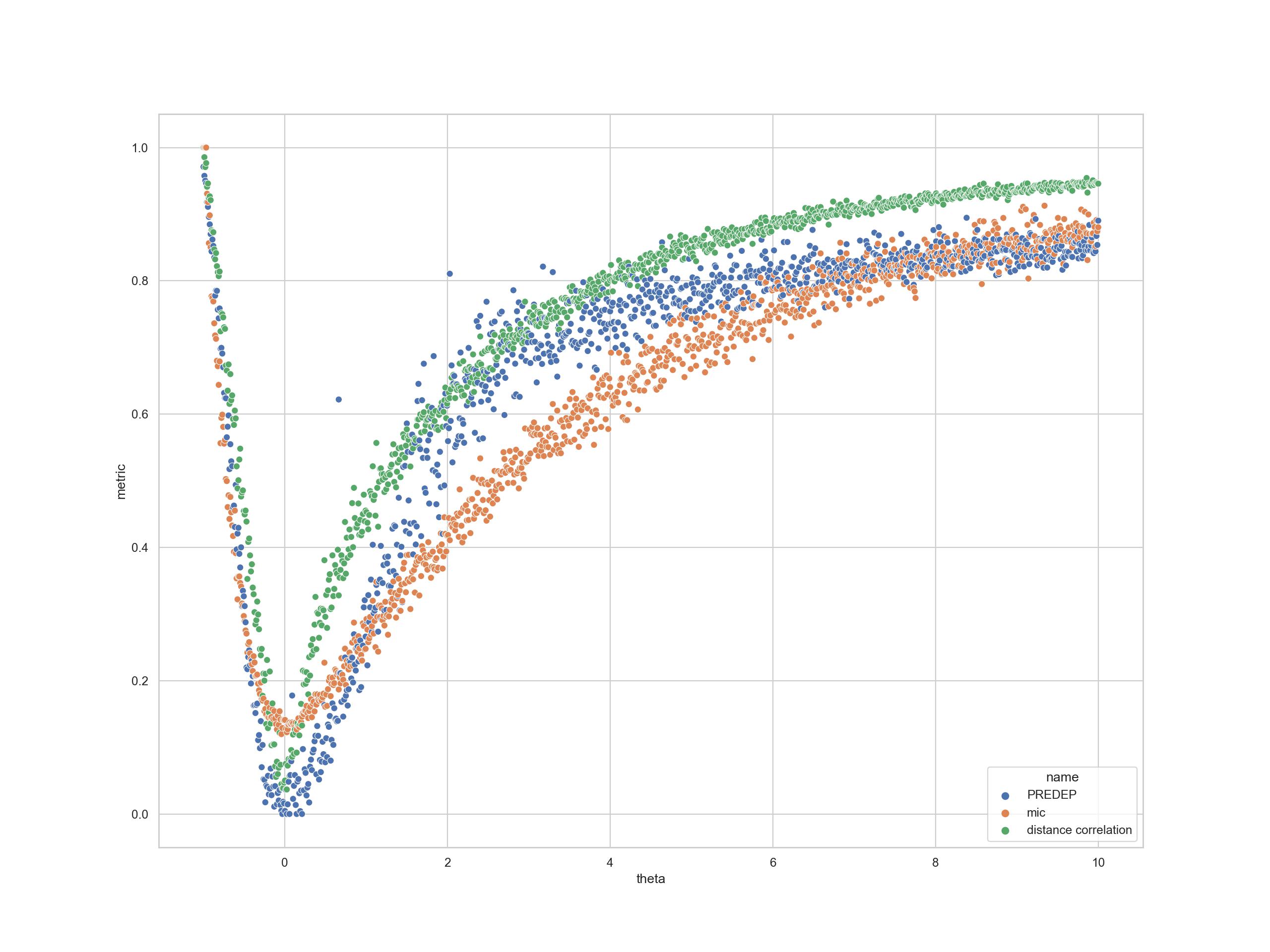}
    \caption{Behavior of PREDEP, MIC, and the distance correlation in synthetic data generated from Clayton copulas as its correlations parameter varies.}
    \label{fig:copula}
\end{figure}

Lastly, in the synthetic data, we examined the behavior of the three metrics on data generated by copulas, specifically Clayton copulas. Two uniform distributions are taken as input, and the copula parameter is varied. Figure \ref{fig:copula} illustrates the measurement outcomes.

From -1 to 0, the values tend to be low, suggesting that PREDEP and distance correlation tend to approach 0, while MIC approaches something close to 0.15. From 0 to infinity, they seem to converge towards 1. This trend might be due to a low number of samples, yet it demonstrates that for data generated by this copula with little relation, PREDEP, and distance correlation have a tendency to approach 0 more readily than MIC.

Moreover, certain observed behaviors include the fact that for values between -1 and -0.2, the metrics exhibit very close values. Between 0 and 6, PREDEP does not seem to align with either of the two metrics' behavior, whereas, beyond 6, it becomes more similar to MIC.

In summary, at the negative extremes, PREDEP behaves similarly to both metrics, particularly to distance correlation. As the theta value becomes high, its behavior appears more akin to MIC. For intermediate values, the metric's behavior deviates from both.

\section{Analysis of the WHO dataset}

\begin{figure}[t!]
    \centering
    \subfloat[][PREDEP vs MIC]{\raisebox{0.15cm}{\includegraphics[width=0.22\textwidth]{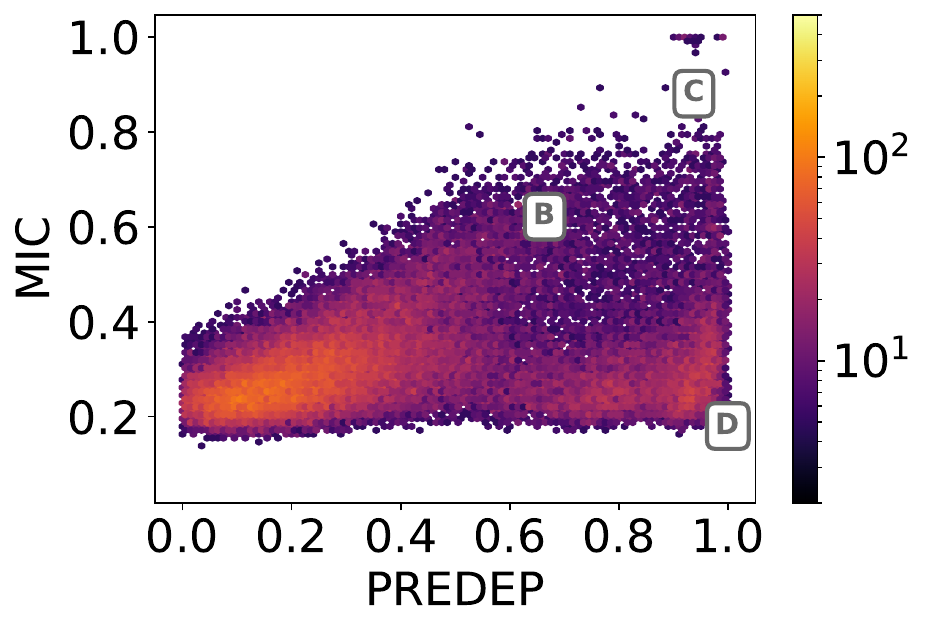}}}\quad
    \subfloat[][Mid MIC/PREDEP \\(B on the scatter)]{\includegraphics[width=0.22\textwidth]{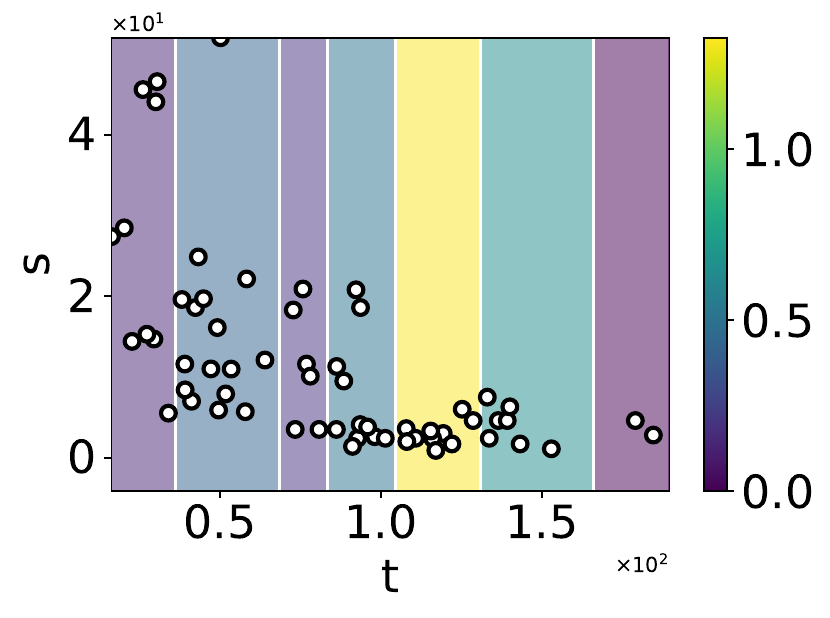}}\quad
    \subfloat[][High PREDEP/MIC \\(C on the scatter)]{\includegraphics[width=0.22\textwidth]{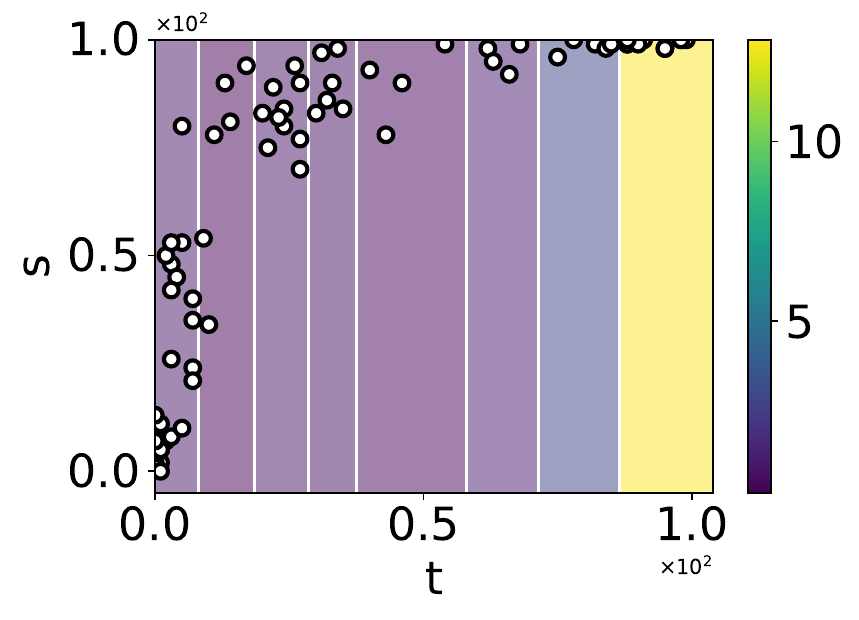}}\quad
    \subfloat[][Low MIC and High PREDEP \\(D on the scatter)]{\includegraphics[width=0.22\textwidth]{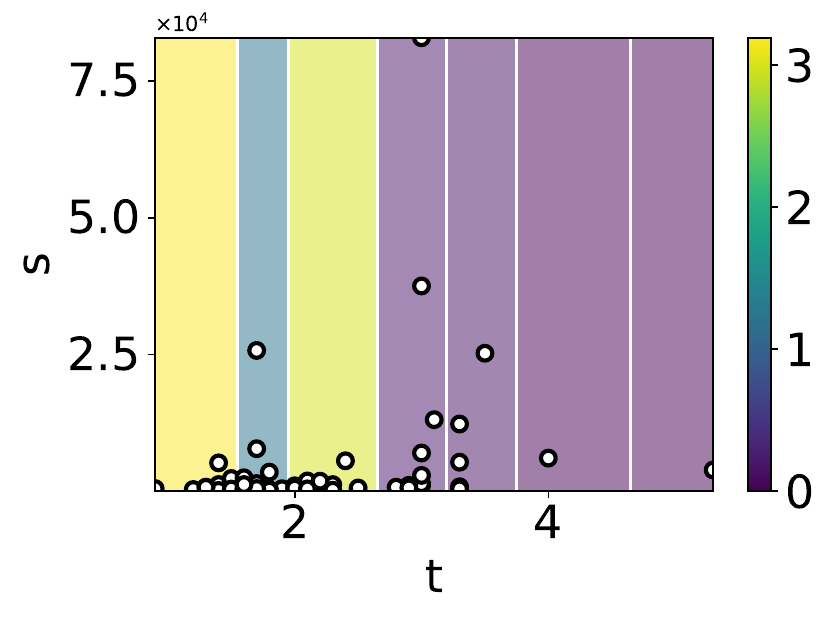}}
    \caption{Application of PREDEP and MIC metrics to the 96,980 selected indicator pairs dataset.}
    \label{fig:who_data_sup}
\end{figure}

In Figure \ref{fig:who_data_sup}, the inset plots correspond to the following pair of variables: 
\begin{itemize}
    \item (B) Scatterplot showing urban Under-5 mortality rate (per 1,000 live births) on the x-axis and lung cancer deaths (per 100,000 men) on the y-axis. 
    \item (C) Scatterplot illustrating the relationship between the percentage of urban population using solid fuels on the x-axis and the percentage of rural population using solid fuels on the y-axis. 
    \item (D) Scatterplot showcasing the connection between under-5 mortality rate (per 1 000 live births) ratio between lowest-highest wealth quintile on the x-axis and number of new female cases of breast cancer on the y-axis.
\end{itemize}

\begin{figure}[t!]
    \centering
    % Primeira linha com dois plots
    \subfloat[][PREDEP vs Pearson Correlation]{\includegraphics[width=0.22\textwidth]{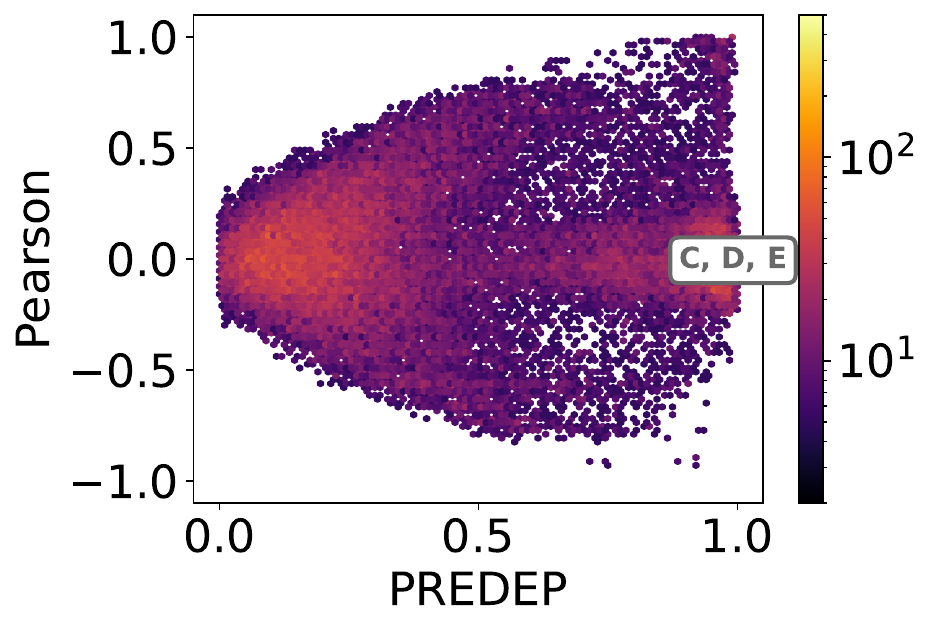}} \quad
    \subfloat[][MIC vs Pearson Correlation]{\includegraphics[width=0.22\textwidth]{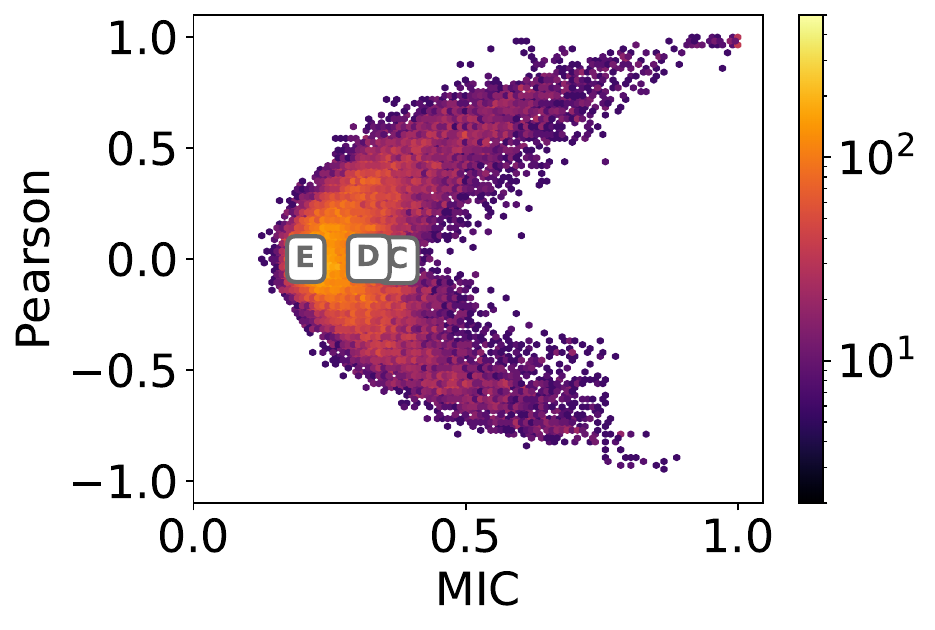}}
    \\
    % Segunda linha com três plots
    \subfloat[][Low Pearson (C above)]{\includegraphics[width=0.22\textwidth]{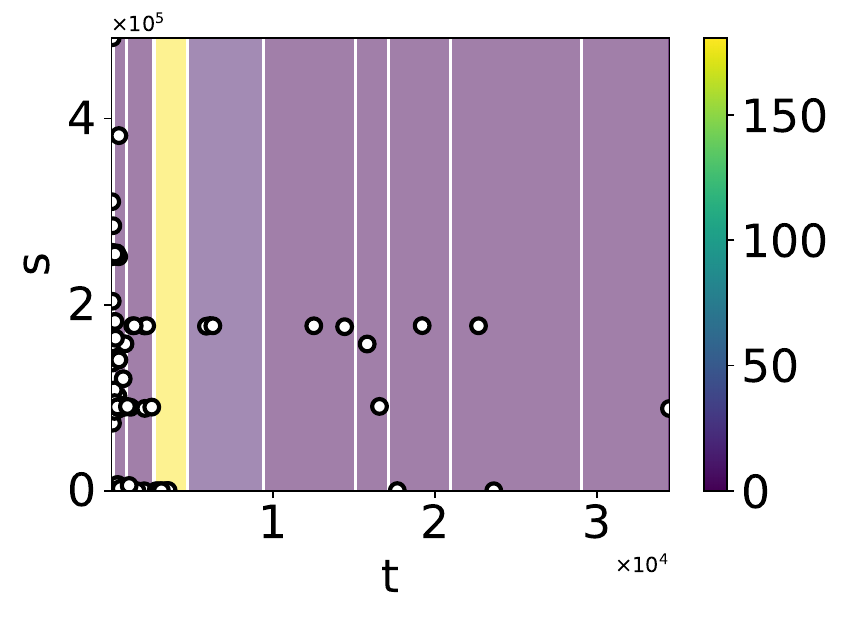}} \quad
    \subfloat[][Low Pearson (D above)]{\includegraphics[width=0.22\textwidth]{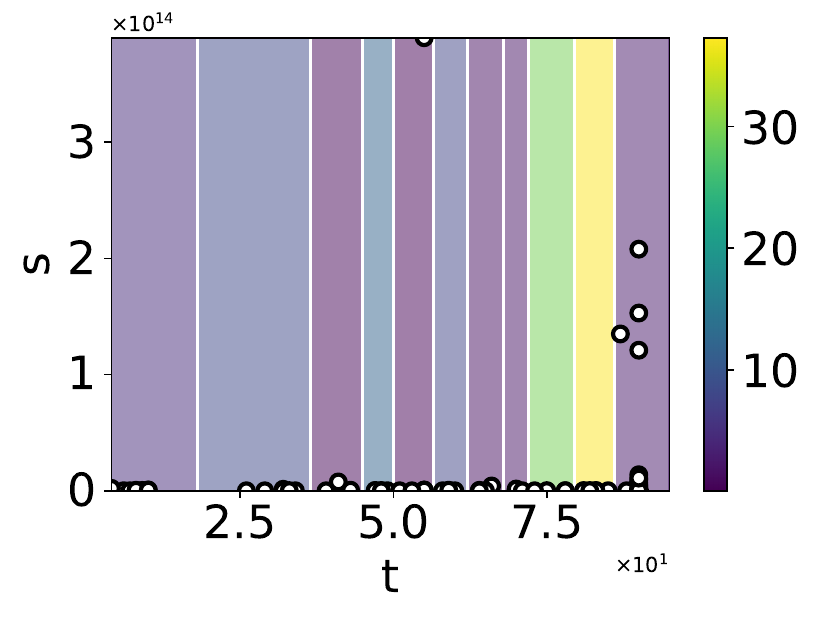}} \quad
    \subfloat[][Low Pearson/MIC (E above)]{\includegraphics[width=0.22\textwidth]{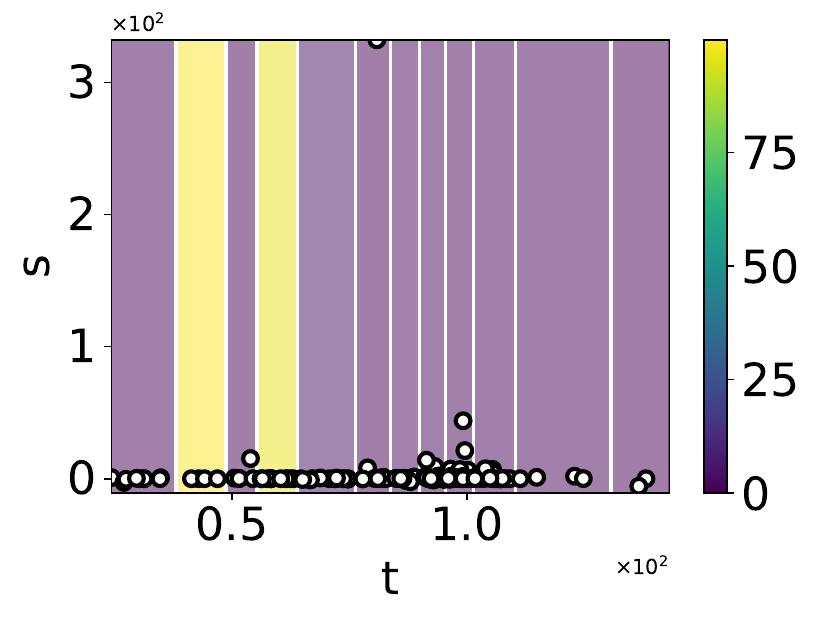}}
    \caption{PREDEP and MIC versus Pearson Correlation on the WHO dataset. On the bottom row, we show three examples where PREDEP is high but the other scores fail to capture dependence.}
    \label{fig:who_data_2_sup}
\end{figure}

In Figure \ref{fig:who_data_2_sup} we show the result of calculating PREDEP, MIC and the Pearson correlation coefficient to the pairs of variables in the WHO dataset.
The variables used in the thumbnails C< D, and E are the following:

\begin{itemize}
    \item (C) Scatterplot displaying the relationship between the prevalence of HIV among adults (per 100,000 population) on the x-axis and patent applications on the y-axis. 
    \item (D) Scatterplot displaying  registration coverage of births (\%) on the x-axis and capital formation on the y-axis. 
    \item (E) Scatterplot illustrating the values of the primary completion rate (total) on the x-axis and foreign direct investment net outflows on the y-axis.
\end{itemize}

The results presented in Figures \ref{fig:who_data_2_sup}A show that there is a relative agreement between the PREDEP and Pearson. However, instances, where PREDEP and Pearson diverge, are observable. 
In Figure \ref{fig:who_data_2_sup}B we have the results for MIC and Pearson. The scatterplot shows the distribution of data points, with a ‘C’ format. 
It's evident that the patterns formed by the data points are distinct in the two comparisons, even though both show that pairs of variables with high Pearson correlation coefficients tend to attain high values of PREDEP and MIC. However, the converse holds true only for MIC, as PREDEP assigns high values to a considerable number of points with Pearson correlation coefficients close to 0.

We examined the distributions of some of these pairs, which can be visualized in Figures \ref{fig:who_data_2_sup}C, \ref{fig:who_data_2_sup}D, and \ref{fig:who_data_2_sup}E. In general terms, these distributions exhibit characteristics akin to that represented in Figure \ref{fig:who_data}D: the $Y$ values are concentrated within a range, with a few exceptions, and this concentration intensifies within the horizontal intervals of lighter colors.

\end{document}